\pdfoutput=1
\documentclass[runningheads]{llncs}

\usepackage{eccv}
\usepackage{eccvabbrv}

\usepackage{graphicx}
\usepackage{tabularx}
\usepackage{booktabs}
\usepackage{multirow}
\usepackage{eucal}
\usepackage{amsfonts}
\usepackage{ bbold }
\usepackage{amsmath}
\usepackage{soul}
\usepackage{lipsum} % For generating dummy text
\usepackage{wrapfig} % For wrapping figures and tables
\usepackage{enumitem}
\usepackage{pgfplots}
\pgfplotsset{compat=1.17}
\usepackage{caption}
\usepackage{booktabs}
\usepackage{algorithm}
\usepackage{algorithmic}
\usepackage{amsmath}
\usepackage{amsfonts}
\usepackage{graphicx}

\newcommand{\benchmark}{Active Memories Benchmark\xspace}
\newcommand{\bs}{AMB\xspace}
\newcommand{\our}{AMEGO\xspace}

\definecolor{darkpastelblue}{rgb}{0.47, 0.62, 0.8}
\definecolor{darkpastelgreen}{rgb}{0.01, 0.75, 0.24}
\definecolor{darkpastelred}{rgb}{0.76, 0.23, 0.13}
\definecolor{orange_assign}{rgb}{1, 0.54, 0.24}
\definecolor{pink_assign}{rgb}{0.82, 0.31, 0.60}
\definecolor{blue_assign}{rgb}{0.20, 0.50, 0.78}
\usepackage{tikz}
\usetikzlibrary{patterns}
\soulregister{\cite}7

\usepackage[accsupp]{axessibility}  % Improves PDF readability for those with disabilities.
\usepackage{hyperref}
\usepackage{orcidlink}

\begin{document}

\title{AMEGO: Active Memory\newline from long EGOcentric videos}
\titlerunning{AMEGO: Active Memory from long EGOcentric videos}

\author{Gabriele Goletto\inst{1}\orcidlink{0000-0002-3778-6460} \quad
Tushar Nagarajan\inst{2}\orcidlink{0000-0002-1627-3842} \quad
Giuseppe Averta\inst{1}\orcidlink{0000-0003-1212-3465} \\
Dima Damen\inst{3}\orcidlink{0000-0001-8804-6238}}

\authorrunning{G.~Goletto et al.}

\institute{$^1$ Politecnico di Torino, Italy \quad
$^2$ FAIR, Meta \quad
$^3$University of Bristol, UK\\ \vspace{6pt}
\url{https://gabrielegoletto.github.io/AMEGO/} \vspace*{-6pt}}

\maketitle
\begin{abstract}
  Egocentric videos provide a unique perspective into individuals' daily experiences, yet their unstructured nature presents challenges for perception. In this paper, we introduce \our, a novel approach aimed at enhancing the comprehension of very-long egocentric videos. Inspired by the human's ability to maintain information from a single watching, \our focuses on constructing a self-contained representations from one egocentric video, capturing key locations and object interactions. This representation is semantic-free and facilitates multiple queries without the need to reprocess the entire visual content. Additionally, to evaluate our understanding of very-long egocentric videos, we introduce the new \benchmark (\bs), composed of more than 20K of highly challenging visual queries from EPIC-KITCHENS. These queries cover different levels of video reasoning (sequencing, concurrency and temporal grounding) to assess detailed video understanding capabilities. We showcase improved performance of \our on \bs, surpassing other video QA baselines by a substantial margin.
\end{abstract}

\section{Introduction}
\vspace{-0.3cm}
Episodic memory is a fundamental aspect of human cognition, which allows us to remember and recall our unique personal experiences \cite{tulving1972episodic}. Recently, there has been growing interest in leveraging first-person or \emph{egocentric} videos to develop artificial episodic memory systems \cite{grauman2022ego4d} that identify temporal segments from the video that contain answers to questions \cite{ramakrishnan2023naq} or occurrences of objects \cite{mai2023egoloc, xu2023my} and activities \cite{mavroudi2023learning,zhang2022actionformer}.

Critically, these approaches build representations of long videos from uniformly sampled frame or clip features, and then train a model to retrieve salient moments from the video using them. This has three drawbacks: (1) they are human activity agnostic --- the simplistic uniform sampling of frames is done without an understanding of where the camera-wearer is, when object interactions occur, or what hand the camera-wearer uses, which are key parameters of human activity, (2) they rely on semantically labelled training data  --- explicitly training encoders to relate the query to the input video representations, and (3) they are difficult to interpret --- the implicit representations do not directly reveal human activity leaving such approaches largely as \emph{black boxes}.

\begin{figure}[ht]
\centering
\includegraphics[width=\linewidth]{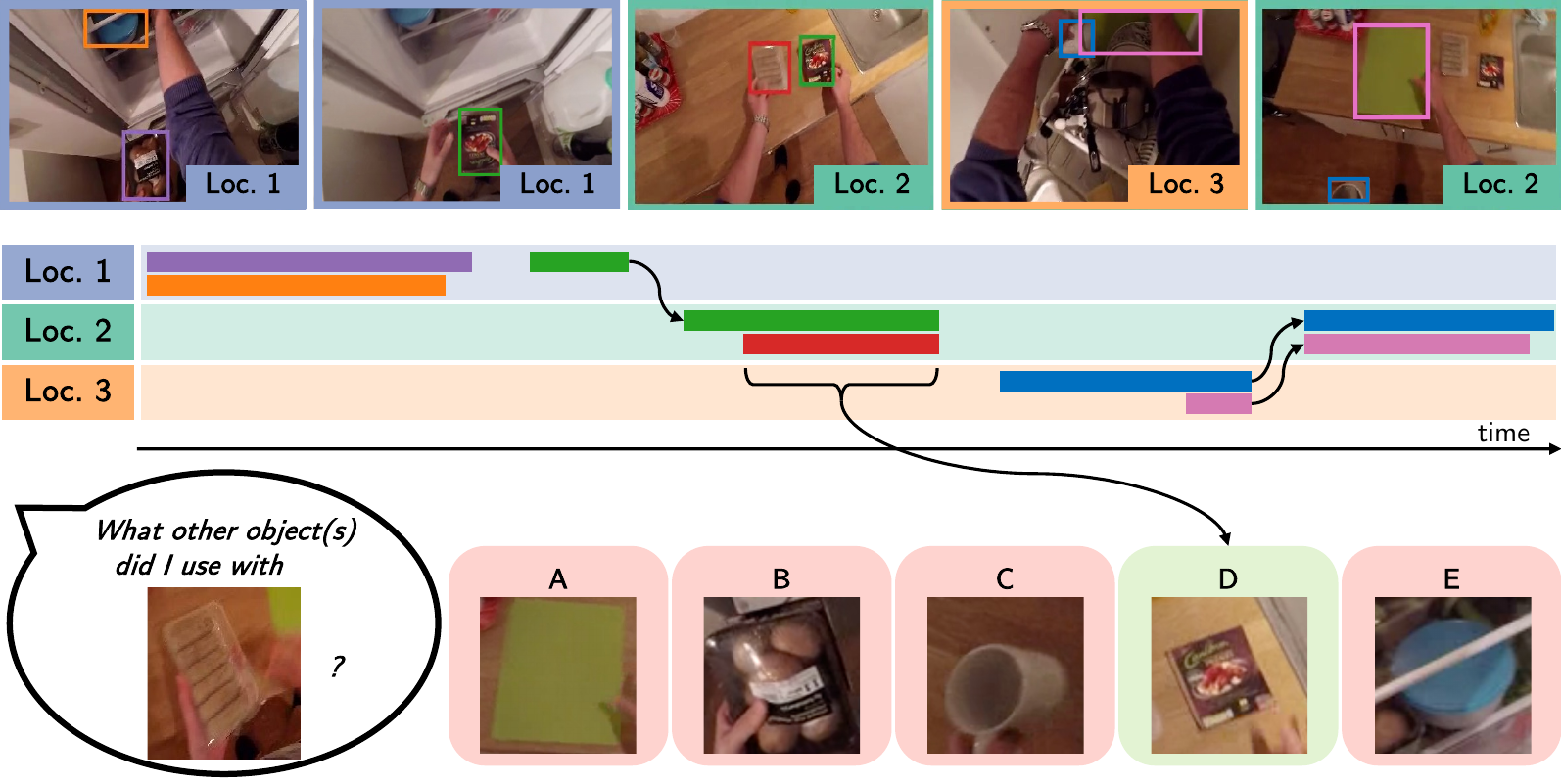}
\caption{
    \our captures key locations and object interactions in a structured representation. In the each frame on top, the external border colour refers to a specific location in \our while colours of objects define specific instances. \our unlocks fine-grained long video understanding allowing multiple queries, such as the one depicted at the bottom of the figure, without reprocessing the long input video.
}
\label{fig:teaser}
\vspace{-0.4cm}

\end{figure}
To address these issues, we present \our, an Active Memory of the EGOcentric video, which serves as an explicit, structured representation that captures both objects interacted with, locations visited, and the interplay between the two (see \cref{fig:teaser}). 
Specifically, \our is composed of (i) a collection of hand-object interaction (HOI) tracklets, which contain consistent interactions between the camera wearer and objects, and (ii) location segments, representing temporal intervals during which the camera wearer engages in activities within specific locations. Importantly, the tracklets and segments are built using visual perception models of human activity and motion rather than a naive sampling of frames. Such models capture information like ``has an object-interaction begun or ended?'', ``is an interaction ongoing, despite hands not visible?'', and so on, leading to representations that are directly tied to activities. 

We populate our \our memory following a three-step process: first, we identify the onset of object interactions; second, we detect the conclusion of ongoing interactions; and finally, we match concluded interactions to previously observed object or location instances. This online pipeline is preferred for efficiently storing and preserving only the relevant information, mirroring the way humans build episodic memories in everyday activities \cite{tulving2002episodic}. Each step uses off-the-shelf visual perception models, resulting in a training-free approach. 

Noticeably, the resulting tracklets and segments are not associated with any fixed taxonomy of objects or locations, resulting in a semantic-free, queryable memory that can be used to answer questions about the video. By simply providing an image of an object or a location of interest, it is possible to access the related information using feature matching. Thanks to its ability to use visual features in a semantic-free manner, \our is more adept at distinguishing objects with subtle differences. This enhances its robustness and flexibility in capturing a wide range of interactions.

To evaluate \our, we propose the \benchmark (\bs) -- composed of more than 20K visual question-answer pairs covering active objects, locations, and their interplay. 
We center the questions around 3 levels of reasoning, i.e. sequencing, concurrency, and temporal grounding. 
Notably, our benchmark is the first to tackle simultaneously interacting objects and locations. \our achieves state of the art results on \bs, surpassing other common Video QA baselines by 12.7\%. This performance underlines its capabilities across the three reasoning levels.
\vspace{-0.3cm}
\section{Related Works}
\textbf{Long video understanding.}
Long videos have gained significant interest recently, largely due to the emergence of large-scale egocentric datasets \cite{grauman2022ego4d, damen2022rescaling, grauman2023ego}. The task involves understanding videos lasting for minutes or even hours, leading to the creation of specialised models \cite{hussein2019videograph, wu2022memvit}. Several question-answering benchmarks have been introduced to evaluate models' proficiency in understanding long videos \cite{tapaswi2016movieqa, yu2019activitynet,lei2018tvqa,yang2021just, fang2024mmbench, du2024towards, li2024videovista, wang2024lvbench, rawal2024cinepile}. Among these, the widely adopted benchmark EgoSchema \cite{mangalam2024egoschema} focuses on videos of up to 3 minutes in duration. Another benchmark, ReST \cite{yang2023relational}, shares similarities with ours as it focuses on visual queries over long egocentric videos.  However, it does not target locations and only emphasises object interactions. Different approaches have tackled long video understanding: some treat it as a natural language question answering task by first captioning the video and then using LLMs to answer queries~\cite{zhang2023simple, wang2023lifelongmemory, ma2024drvideo, park2024too, wang2024videotree, wang2024videoagent}. Others integrate LLMs with a video encoder, leveraging the powerful comprehension and generation capabilities of LLMs~\cite{song2024moviechat, ren2024timechat, qian2024streaming, li2024llms}. Our approach is similar to \cite{fan2024videoagent}, proposing a structured representation of the video, but we specifically focus on interactions rather than indiscriminately memorising all the objects in the video.\\
\textbf{Structured video representations.} Various studies have explored methods for enhancing video representations by incorporating structured information. Contextual relationships have been a key focus, with numerous works investigating relationships between objects and actors \cite{wang2018videos, jain2016structural, baradel2018object, ma2018attend, sun2018actor, cong2021spatial, ji2020action, arnab2021unified}, as well as among actions \cite{brendel2011learning, hussein2019videograph} using graph-based models. In the realm of egocentric vision, efforts have been made to construct structured representations of videos. For instance, \cite{price2022unweavenet} proposes grouping clips by activity threads, while \cite{rodin2023action} introduces egocentric scene graphs to capture interactions of the camera wearer. \cite{nagarajan2020ego}~focuses on constructing a human-centric representation of scenes by capturing the spatial locations of interactions, while \cite{datta2022episodic} builds an allocentric top-down semantic scene representations, grounding the position of objects, from a video capturing a tour of the environment.
Despite addressing various aspects of activities, these approaches do not capture the multiple dimensions inherent in egocentric videos --- namely, object interactions, key locations, and their interplay.\\
\textbf{Video summarisation.} Another related task is video summarisation \cite{meena2023review, rochan2018video, mahasseni2017unsupervised, zhou2018deep} whose aim is to generate a shorter version of the video in the form of key frames or key shots. Egocentric summarisation approaches consider important people and objects \cite{lee2012discovering}, essential events \cite{lu2013story} or aesthetic characteristics of key frames \cite{xiong2014detecting}. Some works have also proposed generating the summaries in an online fashion \cite{lin2015summarizing, zhao2014quasi} but do not target a structured representation of the video.%The work which is more similar
\cite{yagi2021go} proposes a generic object finder, which automatically detects and clusters manipulated objects generating a timeline of the interactions. However, they do not exploit the temporal dimension proposing a system which is affected by noise coming from the detector.
Another work which is related to ours is \cite{xiong2015storyline}. The method introduces a storyline representation for egocentric videos, summarising them based on actors, events, locations, and objects. It allows querying across dimensions using boolean operators. However, it mainly detects predefined attractions and supporting objects, which are visually distinct. Our work focuses on finer activities in cluttered scenes.
\vspace{-0.3cm}
\section{Method - \our}
\vspace{-0.2cm}
Given a long and untrimmed egocentric video, we aim to capture the knowledge of active objects, key locations and their interplay using a unified structured representation. Such a representation must be \emph{self-contained} --- providing a full description of the camera-wearer's interactions with objects and locations --- and \emph{queryable} ---  as it should help retrieve temporal segments in the video indicating when an object was used, when a location was visited and their intersection (i.e. when an object was used in a specific location). In short, it is an Active Memory of the EGOcentric video, named hereinafter \our. 

We decompose the long egocentric video $\mathcal{V}$ into a set of hand-object interaction~(HOI) tracklets ($\mathcal{O}$) and location segments ($\mathcal{L}$). Each \textbf{HOI tracklet} is a spatio-temporal representation of an object consistently interacting with at least one hand of the subject. It is characterised by spatio-temporal bounding boxes and their appearence features. 
Each \textbf{location} segment corresponds to the window of time where the camera-wearer \emph{visits} a location to \emph{perform} interactions, i.e we are interested in activity-centric zones or hot-spots for interactions. 

Put together, the HOI tracklets and location segments form a memory of what objects the camera-wearer interacts with over time, in which locations, and how those objects are moved around the scene. This memory ${\mathcal{E} = \{\mathcal{O}, \mathcal{L}\}}$ is built online, eliminating the need for reprocessing past visual information, and then queried to answer a variety of questions about objects, locations and their interplay, as our experiments will show. Critically, our representations are \emph{semantic-free} --- they represent instances of objects and locations but are not tied to a fixed taxonomy of labels or a known vocabulary. They are tied to the visuals of objects and not to discrete categories, allowing a more fine-grained distinction.

In the following sections, we describe our pipeline to characterise and store object interactions (\cref{sec:det_obj}), to identify location segments (\cref{sec:det_loc}), and then to put them together to form our \our representation. Finally, we describe how to query it to answer various questions (\cref{sec:query_repr}). 
\vspace{-0.4cm}

\subsection{Object Interactions} \label{sec:det_obj}
We begin by characterising object interactions as HOI tracklets $\mathcal{O}$. Each tracklet $o_i \in \mathcal{O}$ is a tuple $(t_s, t_e, b_t, h, id)$ where $(t_s, t_e)$ are the start and end frames of the interaction, $b_t$ is the sequence of bounding boxes representing the object in each frame, $h$ is the hand side that performs the interaction (i.e., left or right), and $id$ is the object instance associated to the tracklet. 

We iteratively build $\mathcal{O}$. At each frame $\mathcal{V}_t$ we perform 3 main steps: (1) initialise possible new candidate HOI tracklets, (2) update the HOI tracklets that are active (i.e. corresponding to ongoing interactions), and (3) store the ones that terminate in the memory $\mathcal{E}$, and assign their corresponding object instance. \\

\noindent \textbf{Initialisation} 
\begin{figure}[t]
\centering
\includegraphics[width=\linewidth, height=6.5cm]{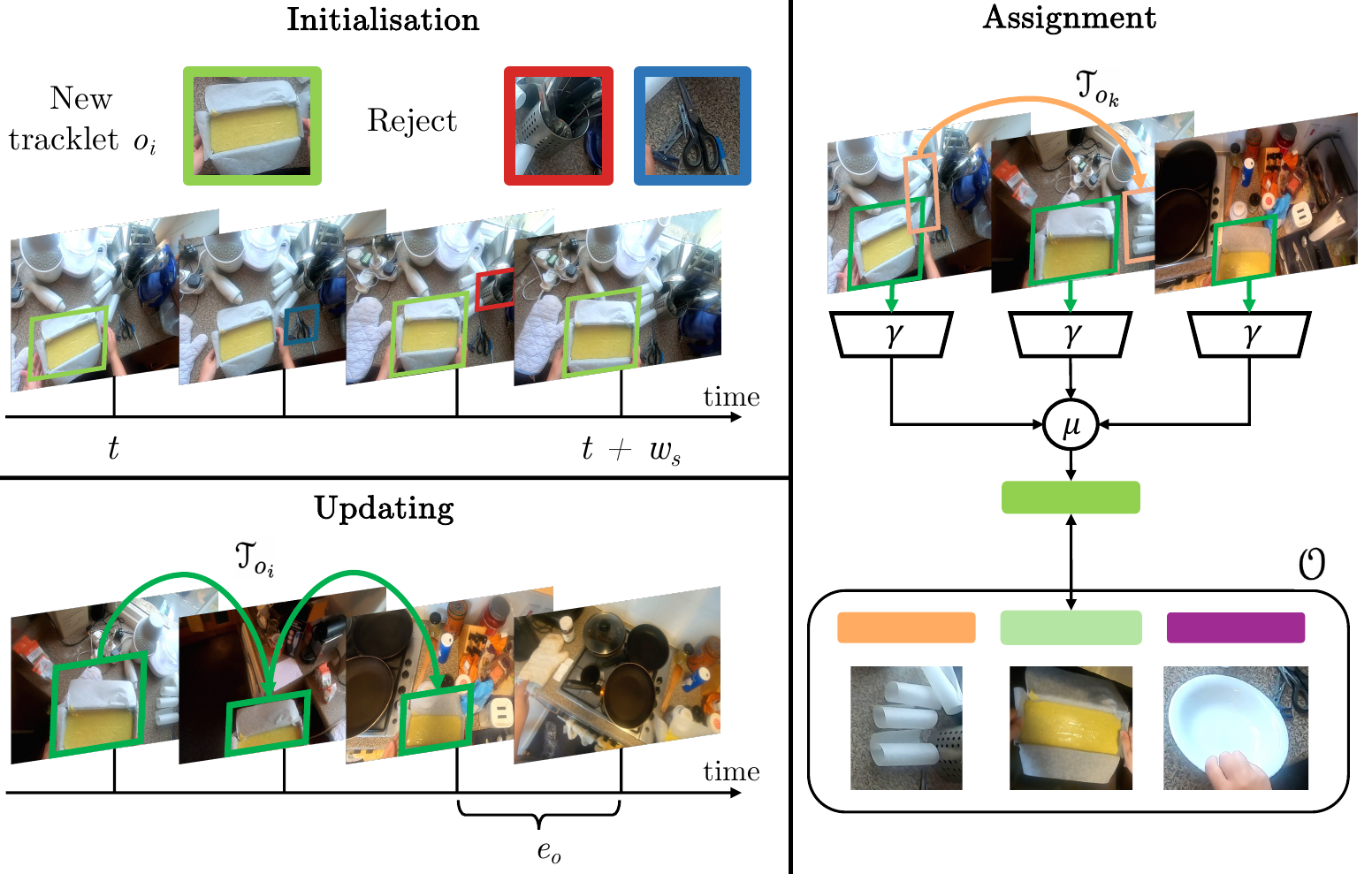}
\vspace{-0.65cm}
\caption{
    We build $\mathcal{O}$ in an online manner, performing the 3 steps depicted at each frame of our video. (i) \textbf{Initialisation} We use consistent active object detections to generate new HOI tracklets. We thus discard noise resulting in sparse detections. (ii)~\textbf{Updating} Once a new tracklet is initialised, we use a SOT tracker ($\mathcal{T}_{o_i}$) to update its detections even when hands go out of the field of view. We end the tracklet when there are $e_o$ consecutive frames with a free hand or a distinctive new object interaction. (iii) \textbf{Assignment} Once a tracklet terminates, we assign it an object instance based on the similarity between its visual features wrt those in memory $\mathcal{O}$.
}
\label{fig:init_method}
\vspace{-0.65cm}
\end{figure}
We use a class-agnostic hand-object interaction detector \cite{shan2020understanding}, which provides a set of active object and hand bounding boxes denoted as $\mathcal{B}^o_t$ and $\mathcal{B}^h_t$ respectively. 
We initiate a new HOI tracklet $o_i$ for each new hand-object interaction, defined as a tubelet comprising at least $s_o$ bounding boxes exhibiting strong spatial overlap within a temporal window of $w_s$ frames (\cref{fig:init_method}, top-left). This spatio-temporal filtering allows us to account for noise as the result of hand-object detectors applied independently over frames. By leveraging the duration of natural hand-object interactions, we can reliably identify new active HOI tracklets, ensuring spatio-temporal consistency in the detections. The HOI tracklet $o_i$ is now considered \emph{active} and is added to $\mathcal{O}$.

For all subsequent frames, we calculate the intersection over union (IoU) for each object interacting with the same hand side. Matching bounding boxes over a threshold, $\theta$, are assigned to the tracklet $o_i$. 
When bounding boxes cannot be assigned to the tracklet, it is considered complete.

\noindent \textbf{Updating} 
\label{sec:update}
Next, we need to capture the entire duration of the interaction, and concurrently, capture all spatial occurrences of the object, performing interaction-aware tracking. While the frame-level HOI detectors are sufficient to identify new interactions, they are unable to reliably extend tracks over time where hands or objects exit the egocentric field of view. Instead, we use an off-the-shelf single-object tracker (SOT) \cite{tang2024egotracks} which can reliably track the object across the whole interaction.  (\cref{fig:init_method}, bottom-left). 

Specifically, for each active object track $o_i$ we initialise a SOT. We consider the track $o_i$ completed if there are no associated detections $\mathcal{B}^o$ for $e_o$ consecutive frames, while the hand $h$ remains visible.
This is because when the hand is out of view, it is likely to still be holding the object.
This results in a spatio-temporal track $\mathcal{T}_{o_i}$ which tracks the object's position, but lacks information about the interaction itself. 

At this point, $o_i$ contains information about its temporal duration (start and end time) and spatial bounding boxes corresponding to the active object, by combining the strengths of frame-based HOI detection and the SOT.\\ 
\textbf{Assignment and storing}
Finally, we match $o_i$ to already seen object instances in our memory. Specifically, given the set of stored HOI tracklets~$\mathcal{O}_t$ observed so far, and the set of running SOT tracks $\mathcal{T}_t$, we check whether $o_i$ can be matched to an existing object instance or if we need to start a new one. To do this, we first compute the visual features of $o_i$:
\begin{equation}
f(o_i) = \frac{1}{|\mathcal{V}_{o_i}|}\sum_{k \in \mathcal{V}_{o_i}}\gamma(k, b^o_k)
\end{equation}
where $\mathcal{V}_{o_i}$ is the set of frames associated with $o_i$, $b^o_k$ is the detection for frame $k$ and $\gamma$ is a visual feature extractor (in our experiments, DINOv2~\cite{oquab2023dinov2}).
To match $o_i$ with instances in $\mathcal{O}_t$, we use an online clustering approach based on $f(o_i)$. The similarity between $o_i$ and a specific object instance $id_j$ is computed as follows:
\begin{equation}
s(o_i, id_j) = \frac{1}{| \mathcal{O}_t \in id_j|} \sum_{\mathcal{O}_t \in {id_j}} <f_{\mathcal{O}_t}, f_{o_i}>
\label{eq:similarity}
\end{equation}
where $\mathcal{O}_t \in id_j$ are the HOI tracklets belonging to instance $id_j$ and $<.>$ measures the cosine similarity.
We assign $o_i$ to the object instance $id_j^*$ that maximizes Eqn.~\ref{eq:similarity}, and is above a specified threshold, $\theta$. Note that if any tracker in $\mathcal{T}_t$ overlaps significantly with $o_i$, and the tracker confidence is higher than the maximum similarity above, then it is assigned to the tracker's instance. Otherwise, $o_i$ is assigned to the corresponding instance $id_j^*$. If the maximum similarity is below the threshold, a new instance is created for $o_i$ (\cref{fig:init_method}, bottom).

At the end of this stage we associate $f(o_i)$ and the assigned instance to $o_i$, and store it into $\mathcal{E}$. We will refer to this \emph{confirmed} tracklet as $\mathcal{O}_i$. It becomes part of \our and can be consequently used in the querying process. 

\subsection{Location segments}
\label{sec:det_loc}
We define the set of location segments $\mathcal{L}$ as the temporal segments when the subject is carrying out interactions at different activity-centric zones. As a subject may interact with multiple objects simultaneously but can only be present in one hot-spot at a time, each location segment $l_i \in \mathcal{L}$ is modelled as a temporal interval corresponding to the start and end of an interaction. Like object interactions, $\mathcal{L}$ is populated online, and in two steps as follows.\\
\textbf{Temporal segmentation}
Given the egocentric frame $\mathcal{V}_t$ and the hand detections $\mathcal{B}^h_t$, to understand whether the hand is interacting with an object while being in a location, we compute the optical flow between $\mathcal{V}_{t-1}$ and $\mathcal{V}_t$ and check hand presence via $|\mathcal{B}^h_t|>0$. We consider the subject carrying out a task if both optical flow has low norm and there is at least one detected hand. 
We used the criteria discussed above as proxies to determine whether the subject has paused (through low optical flow) and is actively interacting with the scene (through a detected hand).
Similar to the process for determining HOI tracklets, we adopt temporal filtering and consider a location segment, $l_j$, to be active only if these two conditions are verified for a consecutive number of frames, $s_l$. Similarly, we terminate $l_j$ when we observe a consecutive number of frames, $e_l$, with either optical flow norm above the threshold or absent of hand detections. 

At the end of this stage, we have temporally defined $l_j$ but we still need to match it to previous location segments at the same hot-spot.\\
\textbf{Assignment and storing} We utilise a visual feature extractor $\sigma$ for locations, to compute average features for the location segment denoted as $g(l_j)$. 
Next, we calculate similarity scores between the stored location instances and $l_j$ by computing the average cosine similarity. We assign $l_j$ to the instance that maximizes the similarity beyond a specified threshold, $\tau$. If the threshold is not met, a new instance is created.
Finally, we pair $g(l_j)$ and the assigned instance to $l_j$, and store it into $\mathcal{E}$. We will refer to this \textit{confirmed} location segment as $\mathcal{L}_j$.

\subsection{Querying \our representations}
\label{sec:query_repr}
\begin{figure}[t]
\centering
\includegraphics[width=\linewidth]{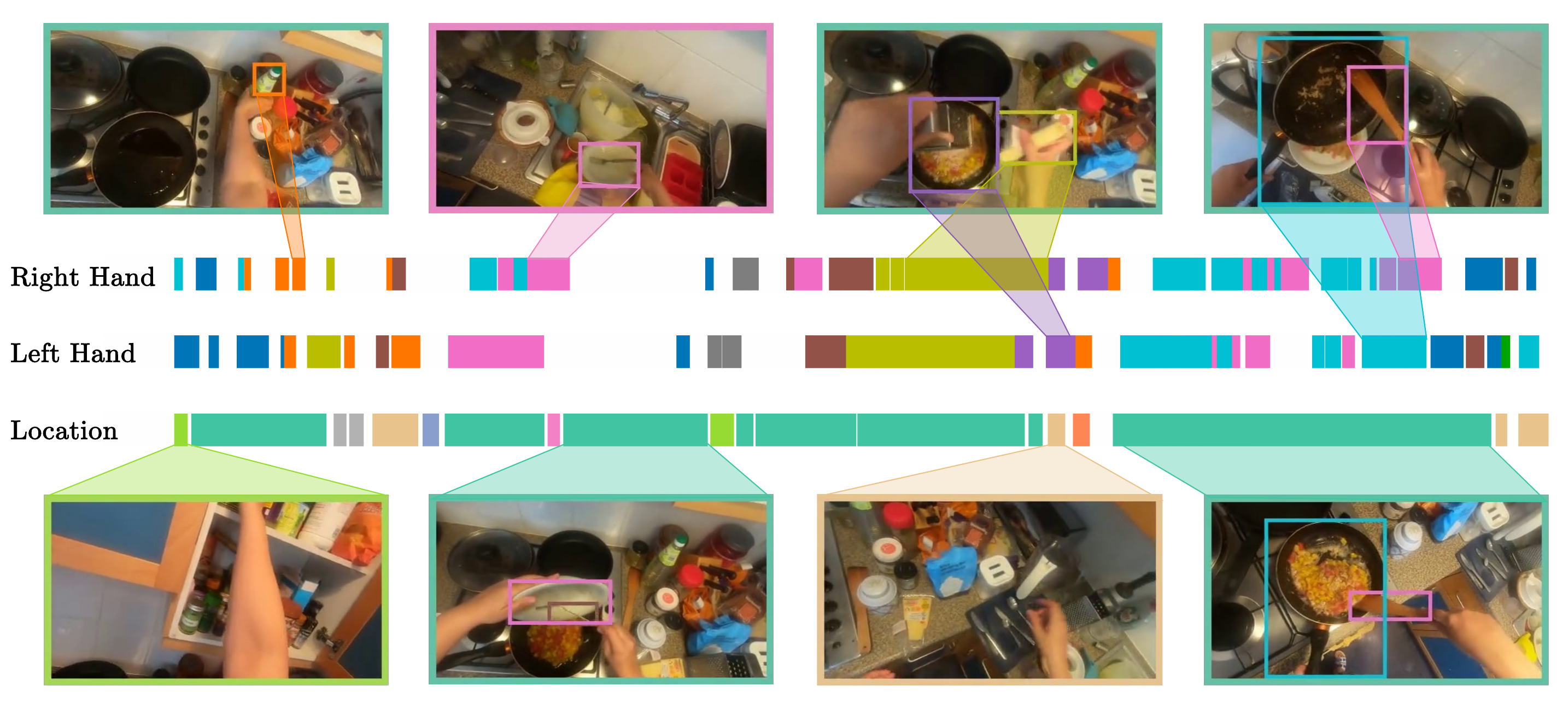}
\vspace{-0.65cm}
\caption{
    An example of \our on a long egocentric video depicting objects interacting with the left and right hand of the subject and the visited locations.
}
\label{fig:AMEGO}
\vspace{-0.65cm}
\end{figure}

After processing the whole video we obtain our \textit{\our}: a complete set of HOI tracklets $\mathcal{O}$ and Location segments $\mathcal{L}$ (see \cref{fig:AMEGO}). Utilising \our, we can determine whether any object has been in use and if the person has interacted at any locations. We achieve this in a semantic-free retrieval manner. Given an object image, $q_o$, we first extract its visual features, $f(q_o)$, using $\gamma$. Subsequently, we assign it an object instance, $q_{id}$, based on the similarity between its visual features and those in $\mathcal{O}$. With the obtained $q_{id}$, we can query $\mathcal{E}$ to retrieve information about its interactions. For instance, by searching for all tracklets in $\mathcal{O} \in q_{id}$, and their interaction intervals $(t_s, t_e)$, we can identify all the temporal segments when $q_{id}$ has been used. 

Similarly, using $\sigma$, we can match any input location image to $\mathcal{L}$ in an identical manner. Consequently, leveraging the common temporal dimension, we can understand where query objects have been used or what objects have been used at the query location. This process allows us to answer any set of queries involving objects and locations without reprocessing the entire video. Inside $\mathcal{E}$, we encapsulate all the information about what occurred in the video. This transforms \our into an active memory of the video that, regardless of queries, is aware of what interactions took place at any point in the video. 
\vspace{-0.3cm}

\section{\benchmark}
\begin{figure}[t]
 \centering
 \includegraphics[width=\linewidth]{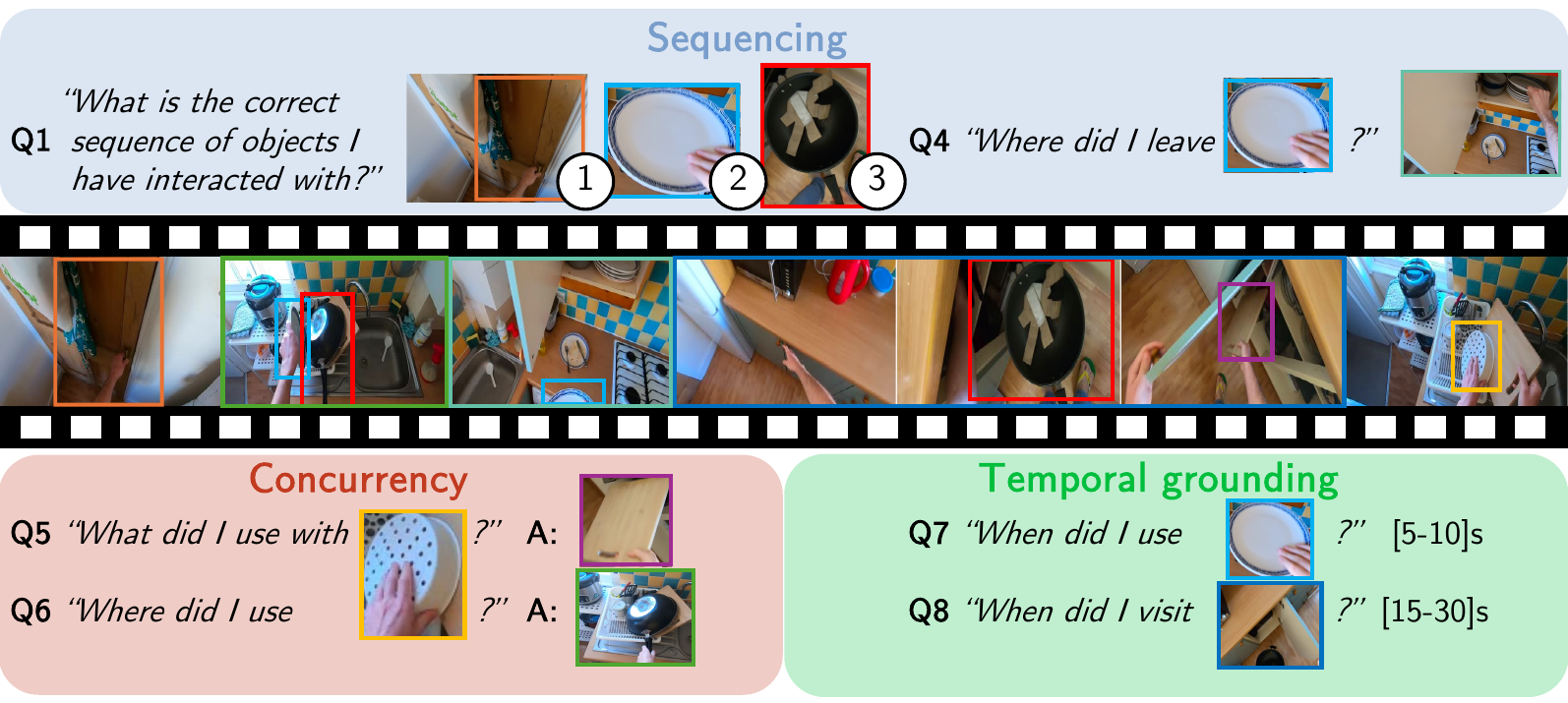}
 \vspace*{-12pt}
\caption{
     Examples queries of \benchmark on an egocentric video (in the middle). We build our benchmark around 3 different levels of reasoning, i.e. \textbf{\textcolor{darkpastelblue}{Sequencing}}, \textbf{\textcolor{darkpastelred}{Concurrency}} and \textbf{\textcolor{darkpastelgreen}{Temporal grounding}}.
 }
 \vspace{-0.3cm}
 \label{fig:bench}
 \end{figure}

We propose the \benchmark (\bs) --- a comprehensive framework to study the interaction between active objects, locations, and their interplay in long egocentric videos, which form key components of daily human activity. 
The benchmark consists of 20.5k queries covering various levels of reasoning. The queries take the form of multiple-choice questions ranging from simple questions about object use (e.g., What did I use with [VQ]? where [VQ] is a visual crop of an object). 
Given a set of [VA] visual answers, the task is to select the correct representation of an object that has been used at the same time as the object represented in [VQ].
Similarly, questions can be answered on the interplay between locations and objects, e.g.
What locations did I use [VQ] in? The answer here would be a set of correct location representations [\{LA, ...\}].
Critically, each visual query of an object [VQ], visual object answer [VA], location query [LQ] or location answer [LA] are parameterised as \emph{visual crops}~\cite{yang2023relational, zellers2019recognition, grauman2022ego4d} to
mitigate the need for a fixed vocabulary or taxonomy resulting in biases associated with language.
Forming a language-free benchmark avoids models that neglect visual data when answering the questions~\cite{yang2020gives, mangalam2024egoschema, jasani2019we}. See Fig.~\ref{fig:bench} for a visual example.

\subsection{Query Criteria}

To construct our benchmark, we build a set of visual query templates that involve objects, locations, and their natural interplay (See \cref{tab:queries}).
We structure our benchmark evaluation around three main reasoning levels, which serve as essential building blocks to enable higher-level activity understanding.

\begin{table*}[t]
\caption{The question templates proposed in our benchmark, along with the corresponding required reasoning, dimensions, types of answers, and number of questions in \benchmark. \textcolor{darkpastelblue}{\textbf{SQ}}, \textcolor{darkpastelred}{\textbf{CO}}, and \textcolor{darkpastelgreen}{\textbf{TG}} represent sequencing, concurrency, and temporal grounding respectively. [VQ] and [LQ] represent object and location crops, while O and L stand for object and location.}
\vspace*{-0.2cm}
\label{tab:queries}
\resizebox{\textwidth}{!}{
\begin{tabular}{c|c|l|l|l|c}
\toprule
\textbf{Reasoning} & \textbf{Query} & \textbf{Template} & \textbf{Dim.} & \textbf{Answer} & \textbf{Qs} \\ \midrule
\multirow{4}{*}{\textcolor{darkpastelblue}{\textbf{SQ}}} & Q1 & What is the correct sequence of objects I have interacted with? & O & Obj. seqs & 4464 \\
& Q2 & What did I use with the left/right hand \emph{after} [VQ]? & O & Obj. & 3466 \\
& Q3 & What did I use with the left/right hand \emph{before} [VQ]? & O & Obj. & 3466 \\
& Q4 & Where did I take/leave [VQ]? & O, L & Loc. & 1266 \\ \midrule
\multirow{2}{*}{\textcolor{darkpastelred}{\textbf{CO}}} & Q5 & What did I use with [VQ]? & O & Obj. sets & 2105 \\
& Q6 & Where did I use [VQ]? & O, L & Loc. sets & 2320 \\ \midrule
\multirow{2}{*}{\textcolor{darkpastelgreen}{\textbf{TG}}} & Q7 & When did I use [VQ]? & O & Intervals & 2614\\
& Q8 & When did I visit [LQ]? & L & Intervals & 809\\ 
\bottomrule
\end{tabular}
}
\vspace*{-22pt}
\end{table*}

\noindent \textbf{\textcolor{darkpastelblue}{Sequencing (SQ)}} questions assess the ability to discriminate the temporal order of events. For example, can the model order interactions in time and identify 
which object the subject used before or after using another object? These are captured by templates Q1-4.

\noindent \textbf{\textcolor{darkpastelred}{Concurrency (CO)}} questions assess 
the ability to capture multiple interactions happening at the same time. For example, can the model reason about whether different objects have been used together (i.e. object-object concurrency), as well as whether an object interaction took place in a specific location (i.e. object-location concurrency)? 
These are captured by Q5-6.

\noindent \textbf{\textcolor{darkpastelgreen}{Temporal grounding (TG)}} questions assess the model's ability to retrieve all intervals of interactions with an active object or a location within the long video. For example, can the model identify when a given object was used or a location was visited (i.e. the start and end time). These are captured by Q7-8.

These three aspects provide a holistic view of information stored in the active memory when observing the long video, serving as the foundational elements for task understanding and causal inference within procedural egocentric videos. 
\vspace{-0.3cm}

\subsection{Benchmark Construction}
We build our queries adopting the templates listed in \cref{tab:queries} and express them as multiple-choice questions. \\
\textbf{Egocentric Videos} We construct our benchmark using 100 videos sourced from the EPIC-KITCHENS dataset \cite{damen2022rescaling}. This dataset is composed of long, unscripted egocentric recordings of participants performing daily living activities in a kitchen environment. On average, the selected videos are 14 minutes long.
More specifically, 18 videos are shorter than 3 minutes, 35 videos are of medium length (between 3 and 10 minutes), 35 videos are long (10-30 minutes) and 12 are very long (> 30 minutes). 
To define our ground truth, we leverage the publicly available dense camera poses from EPIC Fields \cite{tschernezki2024epic} and active object masks from VISOR \cite{darkhalil2022epic}. 
We segmented videos into activity-centric zones by leveraging camera positions from EPIC Fields to track the subject's attention on the scene. We merged EPIC-KITCHENS actions involving prolonged object usage and aligned VISOR masks with class semantics to obtain object bounding boxes. It is important to note that while we utilised these annotations for benchmark construction, our focus was solely on their application in evaluation. Additional details can be found in Supp.\\
\textbf{Query generation} 
We generate queries in a semi-automatic manner from our templates. 
\bs consists of 20.5K multiple choice question answer pairs. Each question consists of five possible options which are extracted semi-automatically from ground truths, to increase the challenge of these questions. In particular, we select candidate answers differently according to the type of question. For instance, for question Q6, options include locations visited immediately after the subject interacted with a specific object. Similarly, for questions Q3-4, options might include objects used with the opposite hand.
This design makes \bs particularly challenging, demanding a detailed understanding of the events in the long video.
Similarly for Q2-3, the query time $t$ is set such that [VQ] is yet to be used -- requiring the search for the interaction with [VQ] first before finding interactions before/after [VQ].
This systematic approach enabled us to create over 20,500 questions, of which 61.7\% are sequencing questions, 21.6\% concurrency questions, and 16.7\% temporal grounding questions, see \cref{tab:queries}. Our dataset comprises 2614 object instances across videos and 809 activity-centric locations. 
On average, short videos (< 3 mins) contain 62 questions, medium-length videos (3-10 mins) contain 134 questions, and long videos (> 10 mins) have 313 questions.

\section{Experiments}
\subsection{Experimental setup}
\subsubsection{Implementation details}
We use the hand-object interaction detector from \cite{shan2020understanding} for identifying object-hand interactions at the frame level. Visual features of objects are extracted using the DINO-v2 pre-trained model \cite{oquab2023dinov2} ($\gamma$), with resizing to $224\times224$ and evaluation on ViT-S and ViT-L versions. Object tracking during interactions employ the EgoSTARK tracker \cite{tang2024egotracks} and we set $\theta=0.6$, $w_s=30$, $s_o=20$, and $e_o=20$.

For locations we use SWAG \cite{singh2022revisiting}, $\sigma$, as the visual feature extractor, trained for image classification using weak supervision of hashtags. This model is currently state of the art in scene classification on Places-365 \cite{zhou2014learning}. We evaluate on ViT-B and ViT-L versions with frames resized to $384\times384$ and $512\times512$. We estimate optical flow with the Flowformer model \cite{huang2022flowformer}, and use a threshold of $2000$ for the optical flow L2 norm. We set $s_l = e_l = 5$ and $\tau=0.5$.\\
\textbf{Baselines}
Our approach is the first able to create a complete representation of the long video which captures multifaced interacting elements. Prior works in this direction focused just on one specific dimension (e.g. locations \cite{nagarajan2020ego} or activities \cite{price2022unweavenet}). SOT trackers would be able to track the object in the video but this would happen regardless of whether the object was interacted with.  
Consequently, we compare \our against common baselines adopted for video QA on the proposed \benchmark: 
\begin{itemize}[leftmargin=*]
\itemsep0em 
    \item[\textbullet] \textbf{Semantic-free QA (SF-QA)} uses vision-language models, i.e. CLIP~\cite{radford2021learning}, to map the query, the video, and the answers into the same embedding space. This process involves extracting visual features from frames of the long video, query patches, and answers, while textual features are obtained from the question. The query embedding is generated by averaging the features from the video, patches, and question. Then, the similarity between this embedding and all answer embeddings is computed. The answer with the highest similarity score is selected. 
    \item[\textbullet] \textbf{SF-QA (obj)} is a variant of SF-QA, with visual features extracted also from active objects detected by \cite{shan2020understanding}. 
    \item[\textbullet] \textbf{Semantic QA (S-QA)} uses off-the-shelf captioners to generate a semantic summary of our video. We use the egocentric video captioner, i.e. LaViLa \cite{zhao2023learning} at 1 fps, as in~\cite{zhang2023simple}, and an image captioner, BLIP-2 \cite{li2023blip}, for generating captions of both the video and of the query patches. 
    Increasing the captioning rate for video would only introduce redundancy in the resulting textual summary. Then, we input captions into an LLM and prompt it with the question. We use LLaMA-2-7B \cite{touvron2023llama} because of its public availability, ensuring the reproducibility of our results. Due to the limited context window, we uniformly subsample textual summaries when they contain more than 4096 tokens.
    \item[\textbullet] \textbf{Multi-round semantic QA (LLoVi)} \cite{zhang2023simple} is similar to the previous one but it queries the LLM twice. First to summarise the video captions given the question, and then to answer the actual query based on the previously generated summary.
\end{itemize}
We evaluate \our alongside the baselines in a zero-shot setting, measuring the accuracy over the queries provided in \bs.
\vspace{-0.3cm}

\subsection{Standalone performance}
We first evaluate the effectiveness of the different \our components against the ground truth. To do so, we manually annotate temporal interactions of objects and locations for two long videos: a 20-minute video from EPIC-KITCHENS \cite{damen2022rescaling} (which is not included in \bs) and a 10-minute video from Ego4D \cite{grauman2022ego4d}. Our annotations identify 22 distinct location instances and 67 different objects, allowing for a temporal comparison with the capabilities of \our in defining interaction intervals. 
\begin{table*}[t]
\caption{Standalone evaluation for HOI tracklets (left) and location (right) segments}
\centering
\vspace*{-0.2cm}
\resizebox{1\textwidth}{!}{%
\begin{minipage}{0.65\textwidth}
\centering
\captionsetup{width=0.9\linewidth}
\label{tab:ablation_hoi}
\begin{tabular}{l|c|c|c}
\toprule
 & AIoU P $\uparrow$& AIoU GT $\uparrow$ &$\mathrm{\Delta N} \rightarrow 0$ \\ \midrule
$s_o = 1$ & 0.08 & 0.49 & 3249     \\
$e_o = 1$ &  0.13 & 0.34 & 537   \\ 
Track w/o hand detections &  0.19 & 0.38 & 222 \\ 
No tracker &  0.19 & 0.39 & 218 \\\midrule
\text{\our} & 0.20 & 0.41 & 210      \\   
\bottomrule
\end{tabular}
\end{minipage}
\hspace{0.05\textwidth}
\hfill
\begin{minipage}{0.65\textwidth}
\centering
\captionsetup{width=0.9\linewidth}
\label{tab:ablation_loc}
\begin{tabular}{l|c|c|c}
\toprule
  &AIoU P $\uparrow$& AIoU GT $\uparrow$ & $\mathrm{\Delta N} \rightarrow 0$ \\ \midrule
$s_l = e_l = 1$ & 0.14 & 0.48 & 163     \\
No flow filter &  0.35 & 0.35 & -1     \\ 
No hand filter &  0.34 & 0.49 & 27   \\ \midrule
\text{\our} & 0.36 & 0.50 & 44    \\   
\bottomrule
\end{tabular}
\end{minipage}
}
\vspace{-0.45cm}
\end{table*}

We evaluate \our using three metrics: 
(i) AIoU P, Average Intersection over Union between each predicted segment and its best-matching ground truth segment, indicating the precision of the predicted tracklets; (ii) AIoU GT, Average Intersection over Union between each temporal ground truth segment and its best-matching predicted segment, evaluating the recall of AMEGO;  (iii) $\mathrm{\Delta N}$, the difference between the number of predicted and ground truth segments. Performance is improved when this number is closer to 0 $\mathrm{\Delta N} \rightarrow 0$;
\cref{tab:ablation_hoi} show the results for the both object and location interactions. 
In particular, it is possible to see the negative effect of noisy detections either at the beginning ($s_o=1$) or at the end ($e_o=1$) of the HOI tracklet.
While AIoU GT is high for $s_o = 1$, this is due to the large number of segments predicted ($\Delta N$ > 3K).
Without using the hand detections, the tracking is stopped after $e_o$ consecutive missing matches, regardless of hand presence.
Accordingly, leveraging hand detections as a proxy to terminate the tracklet helps in detecting long interactions.  Without using the tracker, the method performs worse as it is unable to track the object when the hands exits the field of view. Similar results for location segments show the importance of the various design decisions. It can be noticed how both flow and hand detection help to detect visited locations and, they are complementary.
\vspace{-0.2cm}

\subsection{Results on \benchmark}
To query \our on \bs, we follow simple processes. As an example, to answer temporal grounding queries (Q7-8), we compare the query patch with instances in $\mathcal{E}$, as explained in \cref{sec:query_repr}, then extract the intervals in $\mathcal{E}$ corresponding to the matched instance. Additional details can be found in the Supplementary material. We report results on \our - S, and \our - L, depending on the size of the visual feature extractors adopted (ViT-S/B vs ViT-L).

\begin{table*}[t]
\vspace{-0.3cm}

\centering
\caption{Accuracy results (\%) over the different queries of \bs. Best in \textbf{bold}.}
\label{tab:results}
\resizebox{0.95\textwidth}{!}{
\begin{tabularx}{\textwidth}{>{\centering\arraybackslash}l|*{4}{>{\centering\arraybackslash}p{0.8cm}}|*{2}{>{\centering\arraybackslash}p{0.8cm}}|*{2}{>{\centering\arraybackslash}p{0.8cm}}|>{\centering\arraybackslash}X}
\toprule
 \multirow{2}{*}{\textbf{Method}} & \multicolumn{4}{c|}{\textcolor{darkpastelblue}{\textbf{SQ}}}   &   \multicolumn{2}{c|}{\textcolor{darkpastelred}{\textbf{CO}}}   &   \multicolumn{2}{c|}{\textcolor{darkpastelgreen}{\textbf{TG}}}  & \multirow{2}{*}{\textbf{Total}}  \\ %\midrule
 & \textbf{Q1} & \textbf{Q2} & \textbf{Q3} & \textbf{Q4} & \textbf{Q5} & \textbf{Q6} & \textbf{Q7} & \textbf{Q8} &  \\ \midrule
Random               & 20.0         & 20.0        & 20.0        & 20.0        & 20.0        & 20.0        & 20.0        & 20.0        & 20.0           \\ \midrule
SF-QA               & 13.7         & 21.6        & 22.5        & 26.8        & 22.1        & 31.9        & 23.7        & 26.2        & 22.0           \\
SF-QA (obj)             & 13.1         & 23.4        & 22.6        & 23.2        & 21.7       & 26.1        & 23.8       & 25.2        & 21.2          \\ \midrule
S-QA (LaViLa) & 20.9          & 20.6        & 21.2        & 24.6        & 24.9        & 27.1        & 21.4        & 22.6        & 22.4           \\
S-QA (BLIP-2) & 23.9 & 22.0 & 22.5 & 23.3 & 27.5 & 27.0 & 20.2 & 24.1 & 23.6      \\
S-QA (LaViLa+BLIP-2) & 22.8 & 22.2 & 21.4 & 22.6 & 25.1 & 26.1 & 21.4 & 24.5 & 22.9      \\ \midrule
LLoVi (LaViLa)      &  21.1 & 20.2 & 20.8 & 21.0 & 21.2 & 20.3 & 20.5 & 21.6 & 20.8    \\
LLoVi (BLIP-2)      & 22.3 & 21.4 & 21.8 & 22.2 & 25.6 &  26.7   & 18.1 & 22.2  & 22.4         \\ 
LLoVi (LaViLa+BLIP-2)      & 22.8 & 21.9 & 21.5 & 24.6 & 25.3 &  26.5   & 18.5 & 19.8  & 22.6         \\ 
\midrule
\text{\our} - S    & 32.0         & 35.1        & 34.8        & 35.8        & 24.7        & 37.8        & 33.6        & 44.3        & 33.8           \\
\text{\our} - L    & \textbf{33.7}         & \textbf{36.3}        & \textbf{37.2}       & \textbf{38.3}        & \textbf{27.6}        & \textbf{44.3}        & \textbf{34.7}        & \textbf{48.9}        & \textbf{36.3}        \\   
\bottomrule
\end{tabularx}
}
\vspace{-0.65cm}
\end{table*}

\cref{tab:results} shows the main results on \benchmark. All the baselines struggle to perform slightly better than random among the five answers. Particularly, it is noticeable that despite reaching high results on high-level understanding datasets \cite{mangalam2024egoschema}, Semantic-QA approaches show limited understanding of fine-grained details on long videos. All the baselines show better results on concurrency-related questions (wrt the other reasoning proposed), which may hint at the fact that they might leverage training patterns, e.g. a pan often used at the cooktop. The semantic-free QA baseline performs the worst, demonstrating that features by themselves, without a proper representation, are not enough. On average, BLIP-2 performs better on object-related queries. Indeed, differently from LaViLa, it has been trained on object-centric datasets and therefore shows superior capability to recognise them. Finally, we observe that multi-stage LLM pipelines, such as \cite{zhang2023simple}, perform worse than standard-QA. This likely depends on the fact that directly processing the textual summary reduces the amount of information at subsequent stages for correctly answering the query. 

\begin{wrapfigure}[10]{l}{0.6\textwidth}
\vspace{-0.5cm}
\centering
\centering

\resizebox{0.6\textwidth}{!}{
\begin{tikzpicture}[thick,scale=1, every node/.style={scale=0.8}]
    \def\seqShort{{18.9, 22.5, 30.8}} 
    \def\seqMedium{{19.7, 21.5, 35.7}}
    \def\seqLong{{19.6, 23.4, 36.2}} 
    \def\coShort{{27.8, 26.6, 36.7}} 
    \def\coMedium{{26.5, 27.2,35.6}} 
    \def\coLong{{27.5,27.3,36.6}} 
    \def\tgShort{{26.6,18.3,36.5}} 
    \def\tgMedium{{24.2,19.2,37.4}} 
    \def\tgLong{{24.1,22.2,38.5}} 
    
    \definecolor{colorA}{RGB}{253, 220, 150}
    \definecolor{colorB}{RGB}{246, 180, 100}
    \definecolor{colorC}{RGB}{250, 140, 160}
    
    \pgfmathsetlengthmacro\barwidth{0.3cm}
    \pgfmathsetlengthmacro\barheight{1}
    
    \foreach \i in {0, 1, 2} {
    \fill[colorA] (\i*4*\barwidth + 0.5*\barwidth, 0) rectangle ++(\barwidth, \seqShort[\i]*\barheight);
    \fill[colorB] (\i*4*\barwidth + 1.5*\barwidth,0) rectangle ++(\barwidth, \seqMedium[\i]*\barheight); 
    \fill[colorC] (\i*4*\barwidth + 2.5*\barwidth, 0) rectangle ++(\barwidth, \seqLong[\i]*\barheight); 
    }
    
    % Draw CO bars
    \foreach \i in {0, 1, 2} {
        \fill[colorA] (\i*4*\barwidth + 13.5*\barwidth, 0) rectangle ++(\barwidth, \coShort[\i]*\barheight); 
        \fill[colorB] (\i*4*\barwidth + 14.5*\barwidth, 0) rectangle ++(\barwidth, \coMedium[\i]*\barheight); 
        \fill[colorC] (\i*4*\barwidth + 15.5*\barwidth, 0) rectangle ++(\barwidth, \coLong[\i]*\barheight); 
    }
    
    % Draw TG bars
    \foreach \i in {0, 1, 2} {
        \fill[colorA] (\i*4*\barwidth + 26.5*\barwidth, 0) rectangle ++(\barwidth, \tgShort[\i]*\barheight); 
        \fill[colorB] (\i*4*\barwidth + 27.5*\barwidth, 0) rectangle ++(\barwidth, \tgMedium[\i]*\barheight); 
        \fill[colorC] (\i*4*\barwidth + 28.5*\barwidth, 0) rectangle ++(\barwidth, \tgLong[\i]*\barheight); 
    }
    
    \foreach \i/\label in {0/\textcolor{darkpastelblue}{\textbf{SQ}}, 13/\textcolor{darkpastelgreen}{\textbf{CO}}, 26/\textcolor{darkpastelred}{\textbf{TG}}} {
        \node at (\i*\barwidth + 6*\barwidth, -0.7) {\label};
    }
    \foreach \i/\label in {-4/SF-QA, 0/S-QA} {
        \node at (\i*\barwidth + 6*\barwidth, -0.3) {\label};
    }
    \foreach \i/\label in {9/SF-QA, 13/S-QA} {
        \node at (\i*\barwidth + 6*\barwidth, -0.3) {\label};
    }
    \foreach \i/\label in {22/SF-QA, 26/S-QA} {
        \node at (\i*\barwidth + 6*\barwidth, -0.3) {\label};
    }
    
    \foreach \i/\label in {4/AMEGO, 17/AMEGO, 30/AMEGO} {
        \node at (\i*\barwidth + 6*\barwidth, -0.27) {\label};
    }
    \foreach \i in {12.5, 25.5} {
        \draw[dashed] (\i*\barwidth, 0) -- (\i*\barwidth, 2); 
    }

    \draw[-latex] (11.5,0) -- (11.5,2.4) node[above] {Accuracy (\%)};
    
    \foreach \y in {10,20,30,40,50} {
        \draw (11.6,\y*0.035) -- (11.5,\y*0.035) node[right] {\y};
    }
    \draw (0.2,0) -- (11.3,0) node[below]{};
    
    \node[draw, rectangle, fill=white] at (4*\barwidth, 2) {\begin{tabular}{l l}
        \textcolor{colorA}{\rule{0.5cm}{0.3cm}} & \footnotesize{Short} \\
        \textcolor{colorB}{\rule{0.5cm}{0.3cm}} & Medium \\
        \textcolor{colorC}{\rule{0.5cm}{0.3cm}} & Long \\
    \end{tabular}};
\end{tikzpicture}
}
\caption{Quantitative results depending on the temporal duration of the queried video.}
\label{fig:temp_res}
\end{wrapfigure}
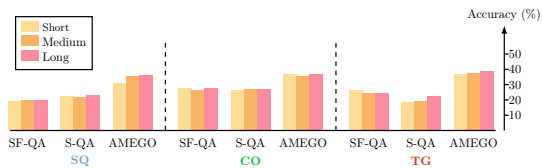
\vspace{-0cm}

\begin{figure}[t]
\centering
\includegraphics[width=\linewidth]{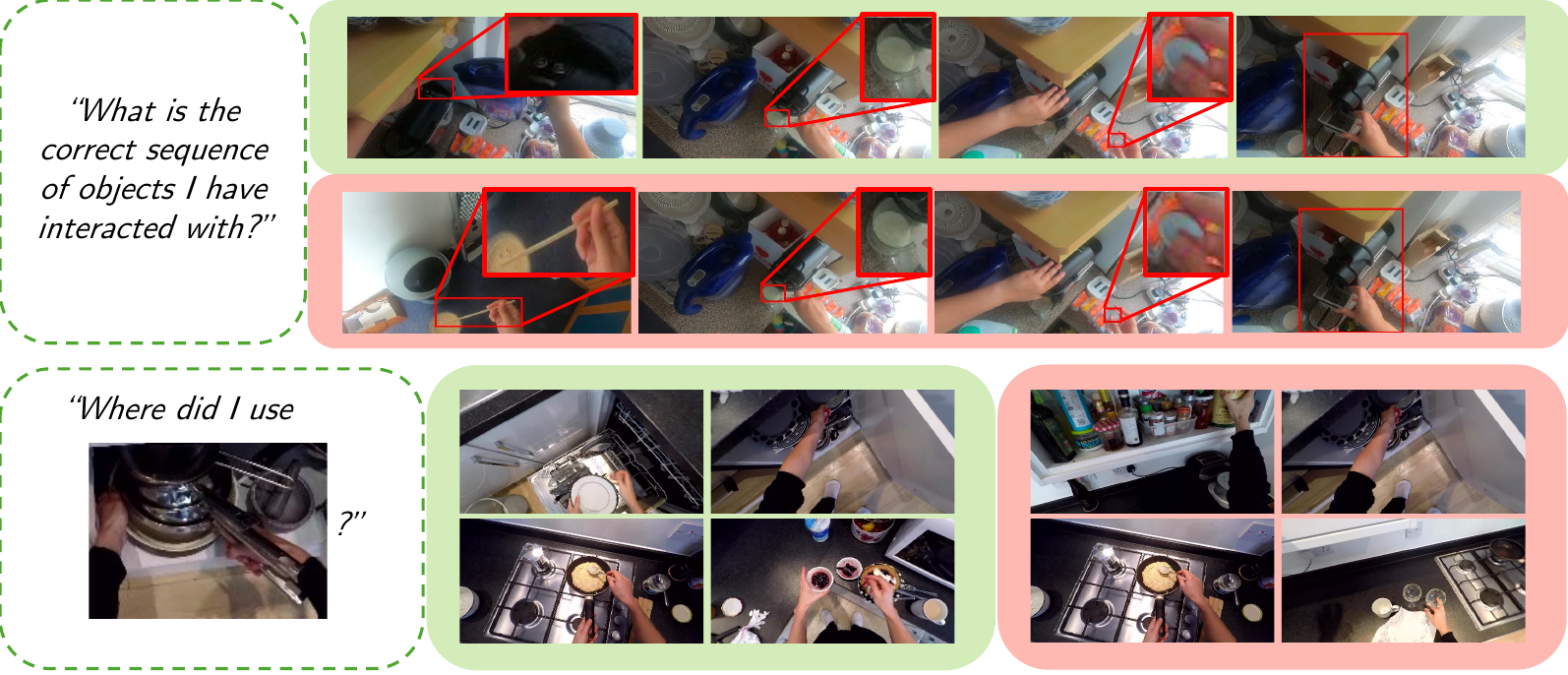}
\caption{
   Qualitative results are presented with sequencing and concurrency queries. Correct answers obtained from querying \our have green background, while incorrect answers from Semantic-QA have red background.
}
\vspace*{-12pt}
\label{fig:qualitative}
\end{figure}

\our achieves good results on the whole set of queries, outperforming baselines by a large margin ($+12.7\%$). It can be noticed that Q5 is the question where \our struggles the most. This difficulty arises from current hand-object interaction detectors facing obstacles in predicting concurrent objects interacting with the same hand of the subject. 
Hence, despite our initialisation and tracking process allowing multiple objects to interact with the same hand, further improvements are needed in this regard.\\
\textbf{Does video duration impact performance?}
We separately evaluate short (<3min), medium (3-10min) and long (>10min) videos. 
\cref{fig:temp_res} compares the best performing semantic-free QA, semantic QA and \text{\our} - L. In general, concurrency and temporal grounding questions are the ones that create more difficulty in long videos for SF-QA and \our. This is reasonable as temporally locating objects and locations in longer videos is intuitively harder.
\\\textbf{Qualitative results}
\cref{fig:qualitative} shows two examples of concurrency and sequencing queries with the answers obtained querying \our (in green) against the ones obtained via Semantic-QA (in red). \our can understand the correct order of usage of items. Indeed it is possible to observe the steps performed by the camera wearer for preparing a coffee (upper part). 
The S-QA approach is not able to capture all fine-grained details in the video and is limited only to part of the sequence.  In the bottom query, instead, it is possible to observe the training biases of LLMs preferring a cupboard (typically used to store a pan) rather than a washing machine.
\vspace{-0.3cm}

\section{Conclusion}
\vspace{-0.3cm}
In this work, we introduce \our, an innovative Active Memory approach tailored for egocentric videos. By dynamically organising interactions and activities into a structured representation, in an online manner, \our mimics the episodic memory cognition. Through semantic-free querying, \our offers a powerful solution for efficient video comprehension without the need for exhaustive reprocessing. 

We evaluate \our on a newly proposed \benchmark which underscores the effectiveness of \our, showcasing its superior performance over common Video QA baselines. This highlights its ability to capture and represent intricate interactions within egocentric videos, paving the way for enhanced video understanding and analysis.\\[3pt]
\footnotesize{\textbf{Acknowledgements} G. Goletto is supported by PON “Ricerca e Innovazione” 2014-2020 – DM 1061/2021 funds and acknowledges travel support from ELISE (GA no 951847).
G. Averta is supported by the project FAIR and Next-GenerationEU (PIANO NAZIONALE DI RIPRESA E RESILIENZA (PNRR) – MISSIONE 4 COMPONENTE 2, INVESTIMENTO 1.3 – D.D. 1555 11/10/2022, PE00000013). This manuscript reflects only the authors’ views and opinions, neither the European Union nor the European Commission can be considered responsible for them. 
D. Damen is supported by EPSRC Visual AI EP/T028572/1 and UMPIRE EP/T004991/1.} 

\footnotesize{We thank Chiara Plizzari for help in extracting the ground truths for \bs, Jian Ma for assistance in computing hand-object detections, and the members of the MaVi group for helpful discussions.}

\bibliographystyle{splncs04}
\bibliography{main}

\newpage
\section*{Appendix}
In this appendix, we report additional details on \our, the \benchmark, additional results and visualisations. 
 In \cref{sec:visualisations}, we include more visualisations of both queries and qualitative results across the complete range of questions in \benchmark (\bs). 
We further detail \bs in \cref{sec:stats}. We then give more information on how we query \our to obtain the answers required for the set of questions in \bs (\cref{sec:query_answer}). Next, in \cref{sec:more_results}, we present additional ablations for \our. Finally, in \cref{sec:pseudocode}, we report the pseudocode version of the pipeline adopted in \our.

\vspace*{-12pt}

\section{Qualitative results}
\label{sec:visualisations}
\begin{figure}[ht!]
\centering
\includegraphics[width=\textwidth]{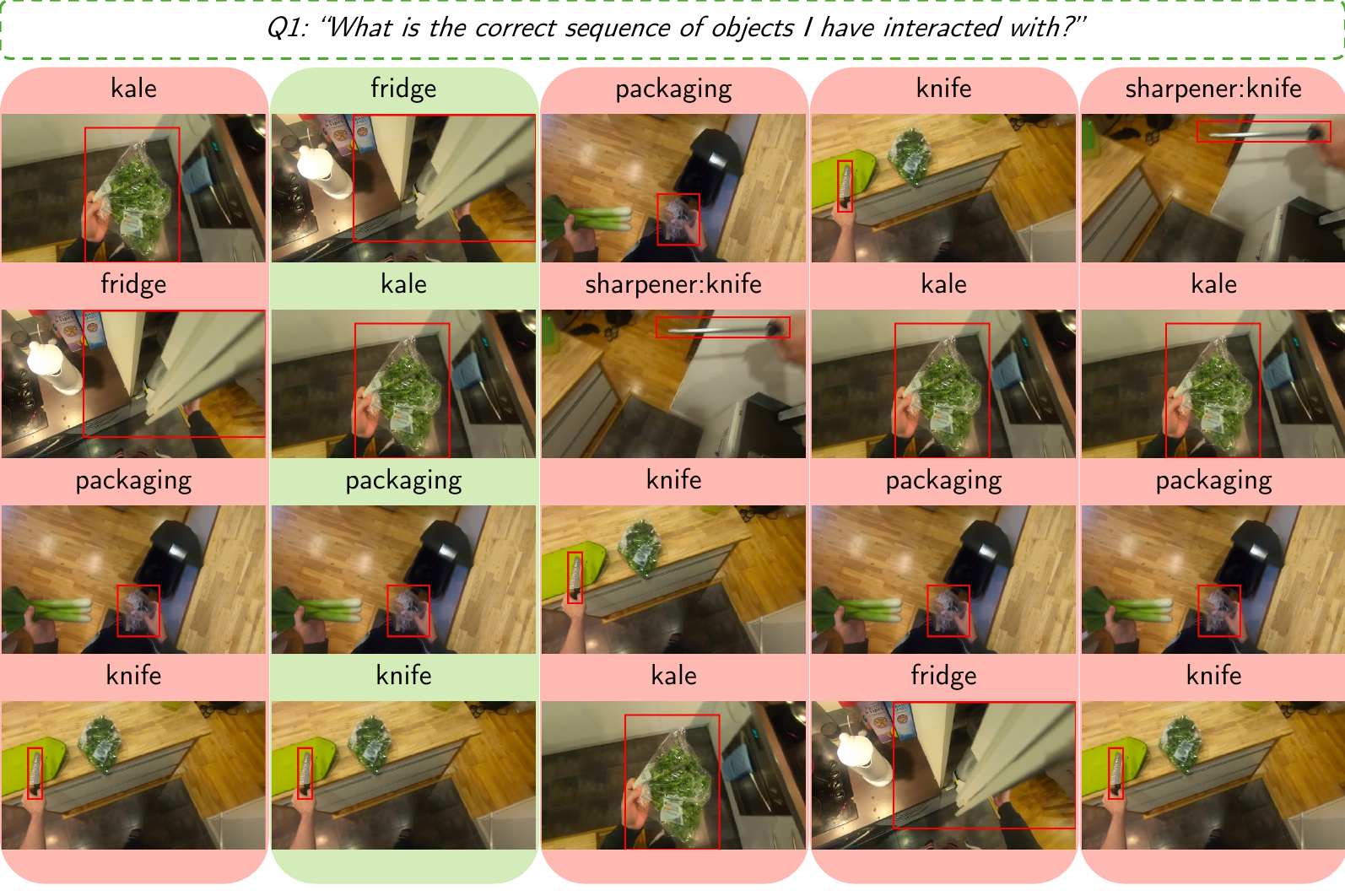}\\
\includegraphics[width=\textwidth]{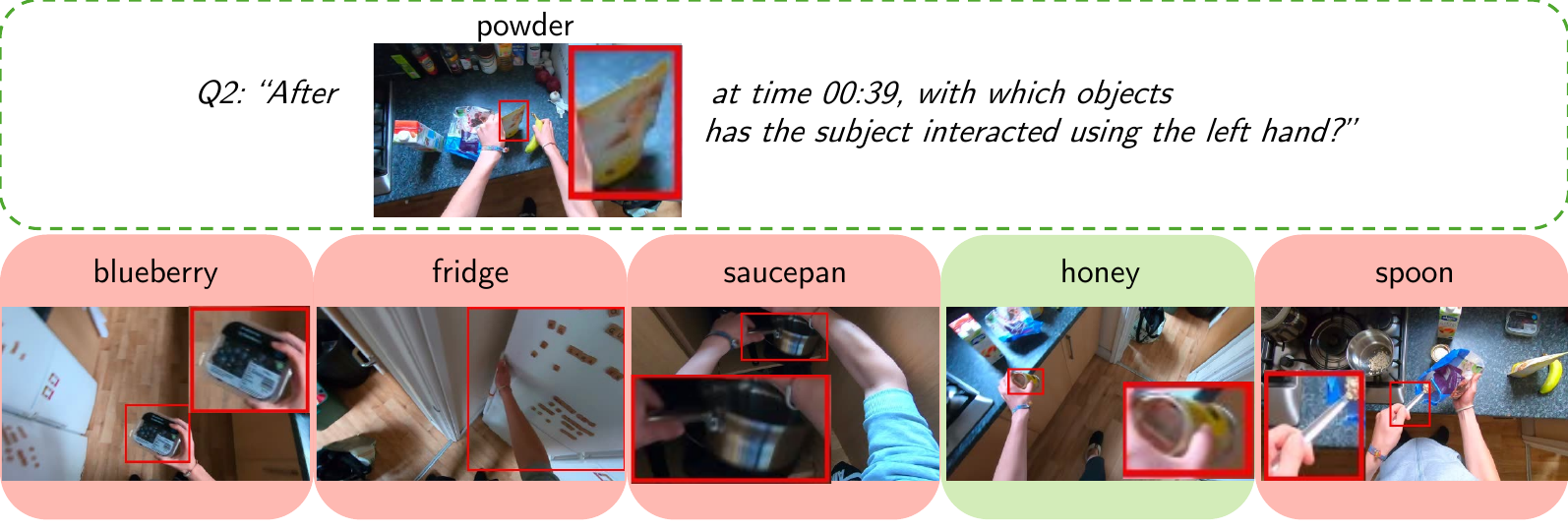}\\
\includegraphics[width=\textwidth]{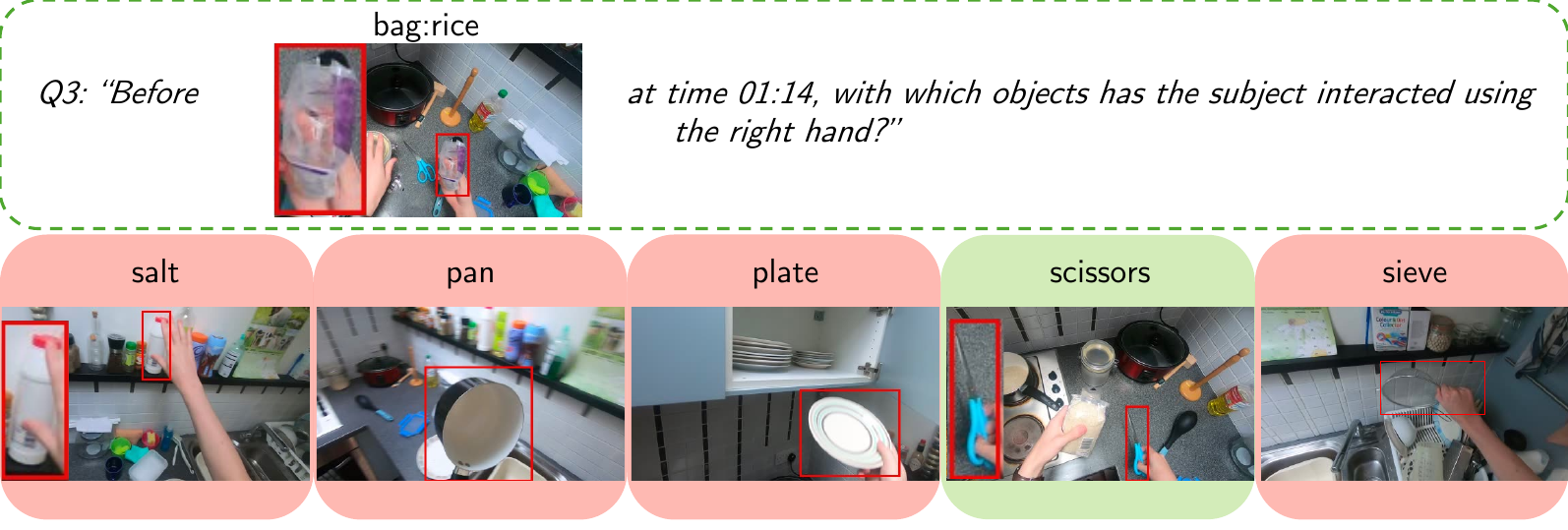}\\
\includegraphics[width=\textwidth]{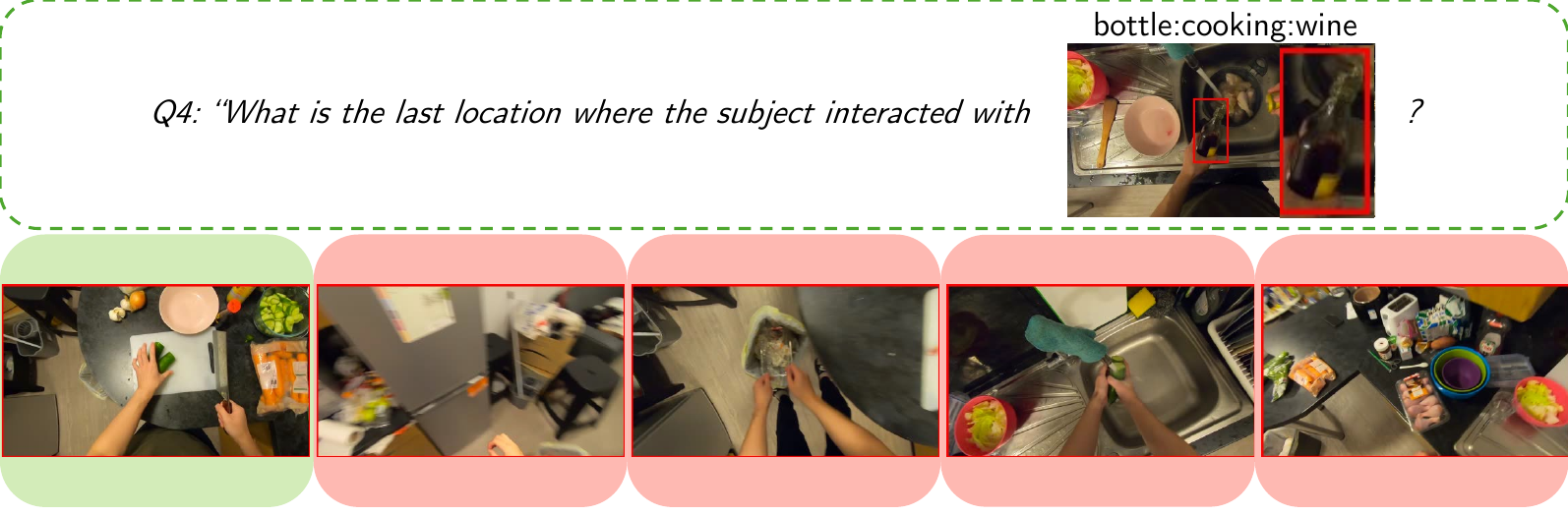}
\caption{
    Examples of Q1-Q4. In green the right answer, correctly selected by \our.
}
\label{fig:qual}
\end{figure}

In \cref{fig:qual} we show additional examples of sequencing questions, with all possible alternatives. \our is able to correctly answer them, showcasing its sequencing capabilities.

On the project webpage we include videos depicting \our representation on EPIC-KITCHENS videos. Similar to Fig. 1 in the main paper each row shows a different location spotted with our method. The white bar represents the temporal position of the frame depicted.

\section{\benchmark}
\label{sec:stats}
\subsection{Ground Truth}
We combine annotations from EPIC-KITCHENS \cite{damen2022rescaling}, VISOR \cite{darkhalil2022epic}, and EPIC Fields \cite{tschernezki2024epic} to extract the ground truth used for creating our queries. 

To obtain ground truths for locations, we filter out all frames with a high optical flow norm as these correspond to segments of video where the camera wearer is moving between locations. We then compute the intersection between the rays tracing from the camera through 5 pixels representing a crop of the image (four corners and centre pixel) and the mesh of the scene.
For a frame size of \( 480 \times 854 \), we selected the following pixels : (213, 240) as central-left, (640, 240) as central-right, (427, 120) as central-top, (427, 360) as central-bottom, and (427, 240) as the central frame. 
We then average the 3D points obtained, representing the locations where the subject focused for a period of time, indicating a potential interaction.
The average is employed to reduce errors arising from noisy automatically extracted meshes.

We then use hierarchical density-based clustering to obtain rough spatial clusters of the scene using the L2 distance among 3D locations as metric. Subsequently, we manually refined the clustering results to segment the videos into different functional activity-centric zones. For example, we differentiating a cooktop from the kitchen counter immediately adjacent to it, as they naturally afford different actions.
We then find temporal segments corresponding to each location cluster.
This approach enabled us to establish ground truths for temporal segments of locations.
\clearpage
To accurately capture object interactions' ground truths, we use action segments from EPIC-KITCHENS where brief interactions with the same object occurred. 
For example, we identify paired actions: `open fridge' - `close fridge' as well as `pick plate' - `put down plate'.
By connecting these actions and finding the temporal extent between them, we can define the full interaction with objects.
This approach allowed us to obtain the complete interaction interval, even when the camera wearer moved objects around the scene. 
We filtered out cases where different instances of the same objects appeared multiple times within a single video. Finally, we adopted VISOR masks to extract the visual queries for \bs.

\subsection{Query creation}
The \benchmark is a visual-only QA benchmark focused on the subject's interactions during long egocentric videos. 
One of the challenges is to select a visual representation for objects to form our visual query [VQ].
To address strong occlusion typical in egocentric vision, caused by the camera wearer's hands or other objects, we selected up to 3 different image patches for each object to form the query. 
These patches should be temporally distinct, to showcase different poses -- we use a minimum of 0.5s between patches.
Additionally, we select patches with minimal spatial overlap with bounding boxes of other active objects/hands in the same frame, to minimise occlusion. Similarly, for location images, we extracted frames with the lowest spatial overlap with active objects, so the location is present without many moving objects. 
For locations, we use location images with a minimum of 1s differences.

\begin{figure}[htbp]
    \centering
    \begin{subfigure}[b]{0.45\textwidth}
        \centering
        \includegraphics[width=\textwidth]{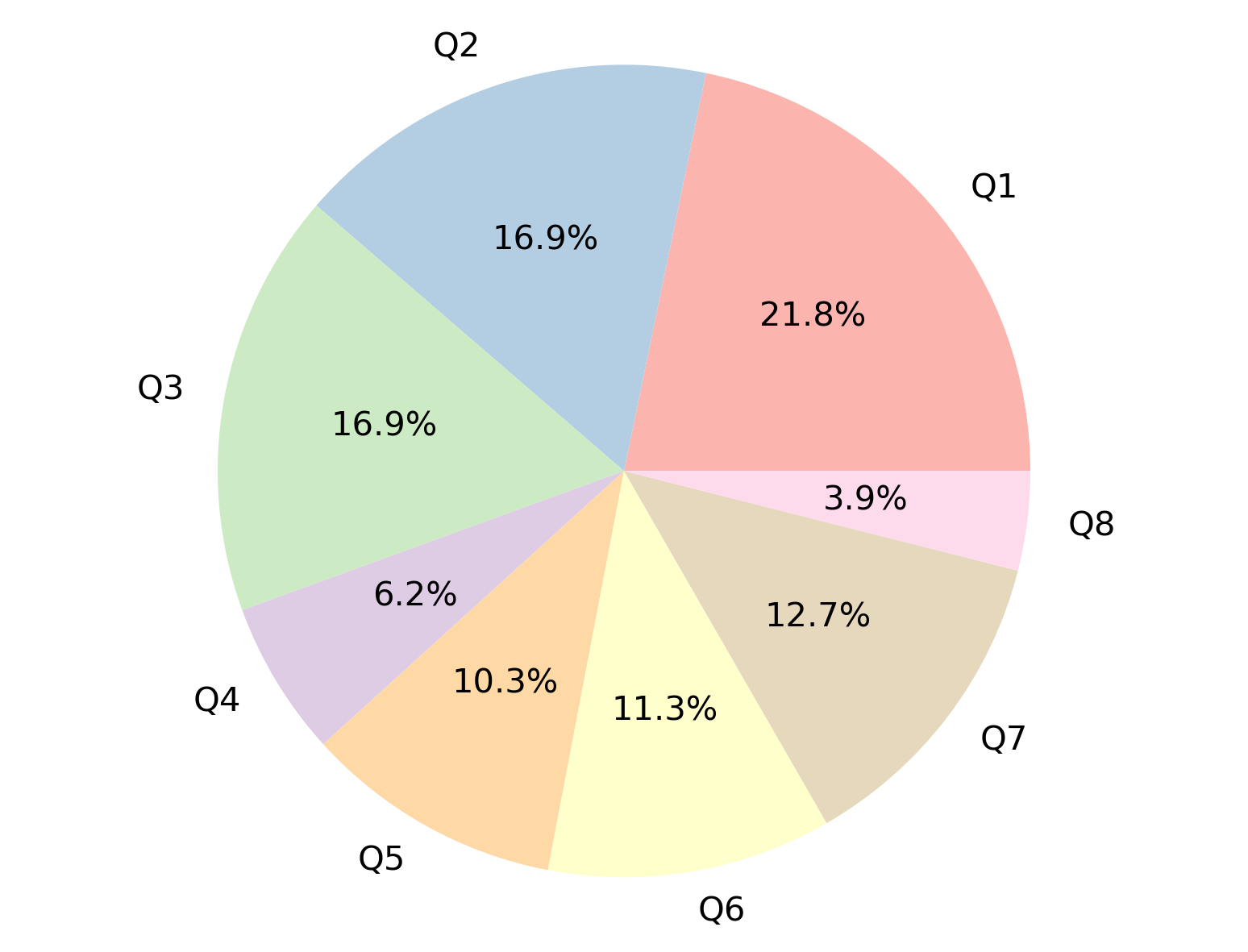}
        \caption{Distribution of queries by type}
        \label{fig:sub1}
    \end{subfigure}
    \hfill
    \begin{subfigure}[b]{0.48\textwidth}
        \centering
        \includegraphics[width=\textwidth]{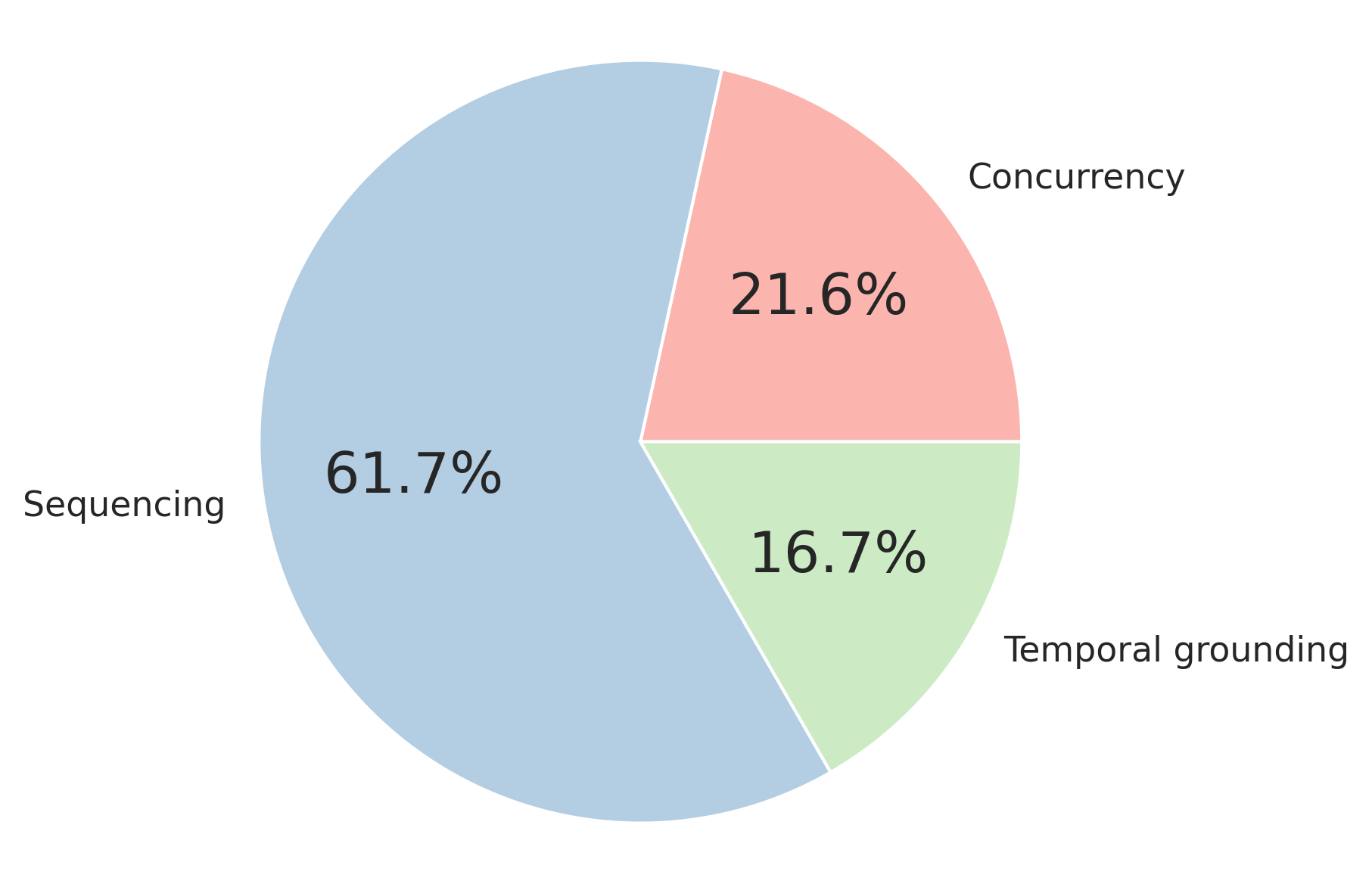}
        \caption{Distribution of queries by reasoning level}
        \label{fig:sub2}
    \end{subfigure}
    \vskip\baselineskip
    \begin{subfigure}[b]{0.48\textwidth}
        \centering
        \includegraphics[width=\textwidth]{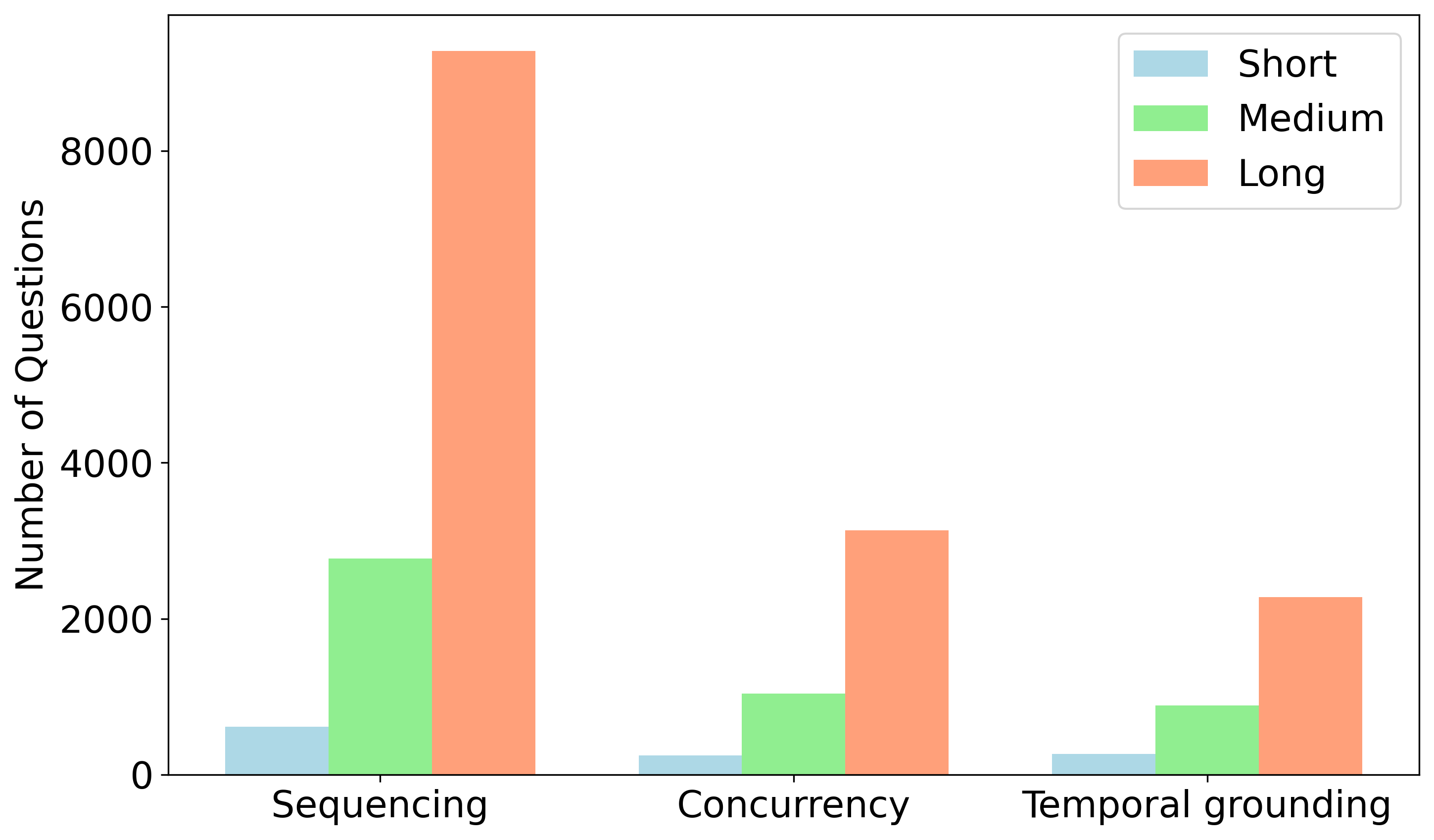}
        \caption{Distribution of queries by video duration for each reasoning level}
        \label{fig:sub3}
    \end{subfigure}
    \hfill
    \begin{subfigure}[b]{0.48\textwidth}
        \centering
        \includegraphics[width=\textwidth]{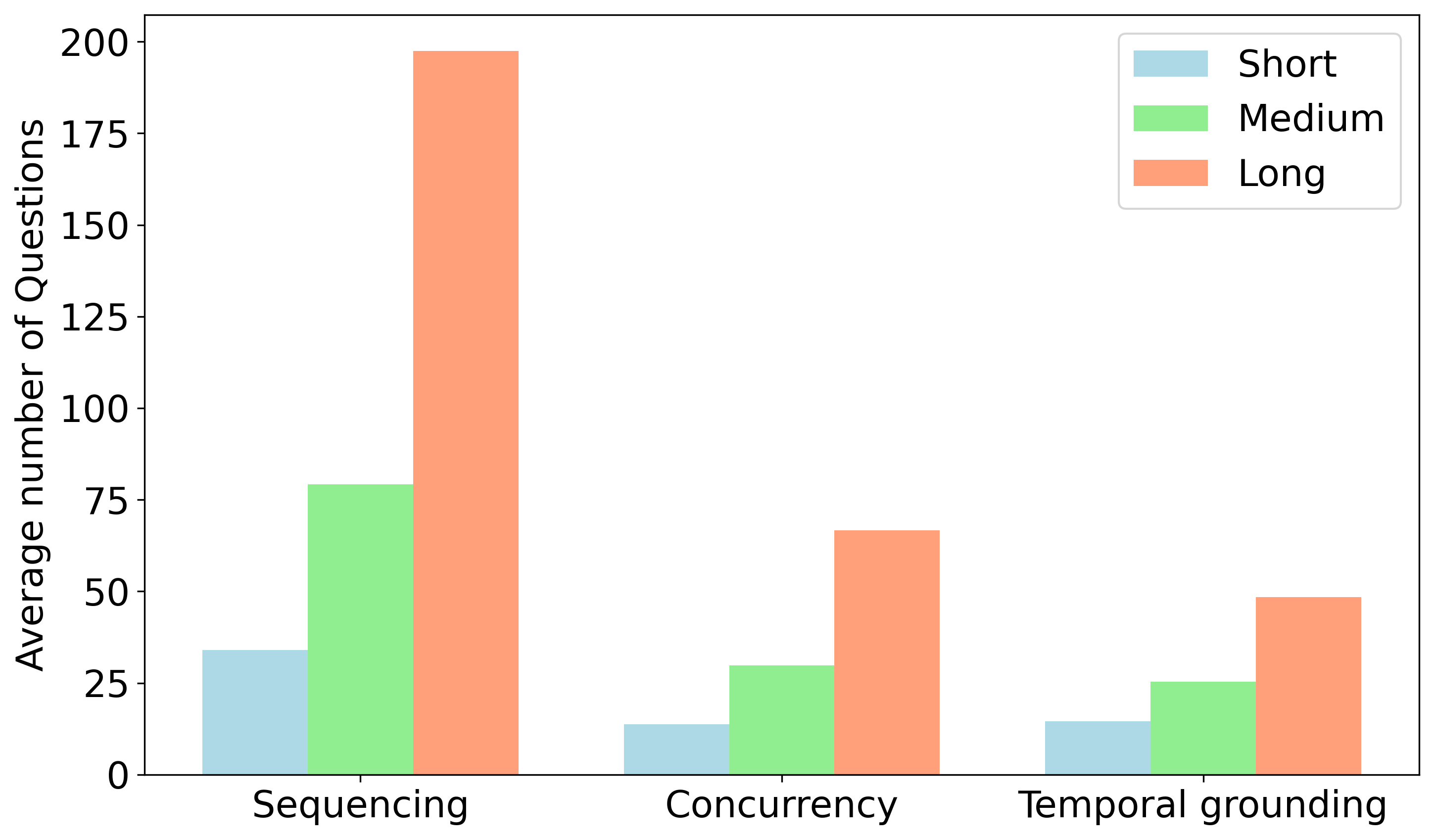}
        \caption{Average number of queries per video by duration for each reasoning level}
        \label{fig:sub4}
    \end{subfigure}
    \vskip\baselineskip
    \begin{subfigure}[b]{0.48\textwidth}
        \centering
        \includegraphics[width=\textwidth]{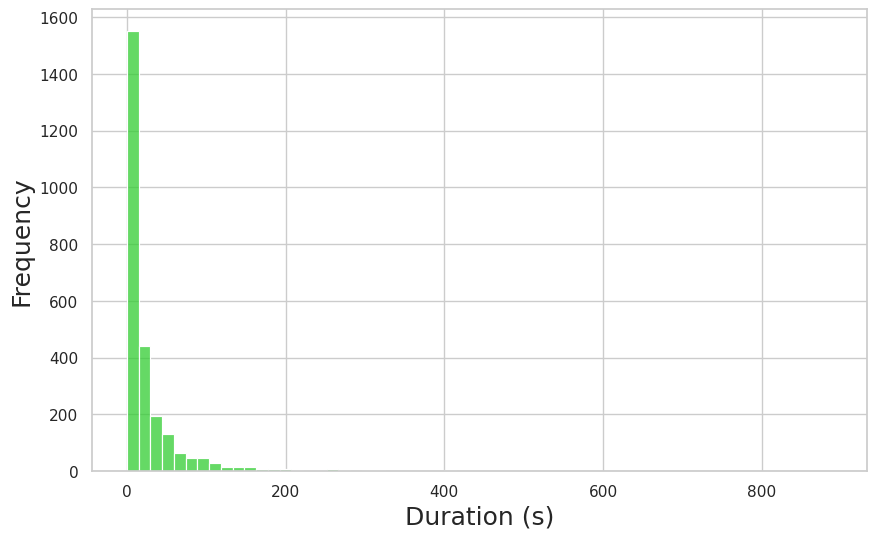}
        \caption{Queries of type Q7 by duration}
        \label{fig:sub5}
    \end{subfigure}
    \hfill
    \begin{subfigure}[b]{0.48\textwidth}
        \centering
        \includegraphics[width=\textwidth]{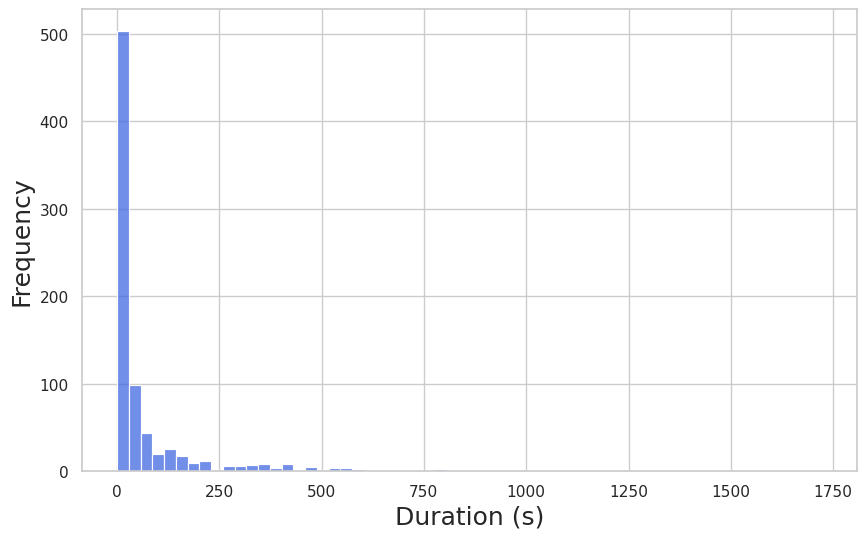}
        \caption{Queries of type Q8 by duration}
        \label{fig:sub6}
    \end{subfigure}
    \vskip\baselineskip
    \begin{subfigure}[b]{0.48\textwidth}
        \centering
        \includegraphics[width=\textwidth]{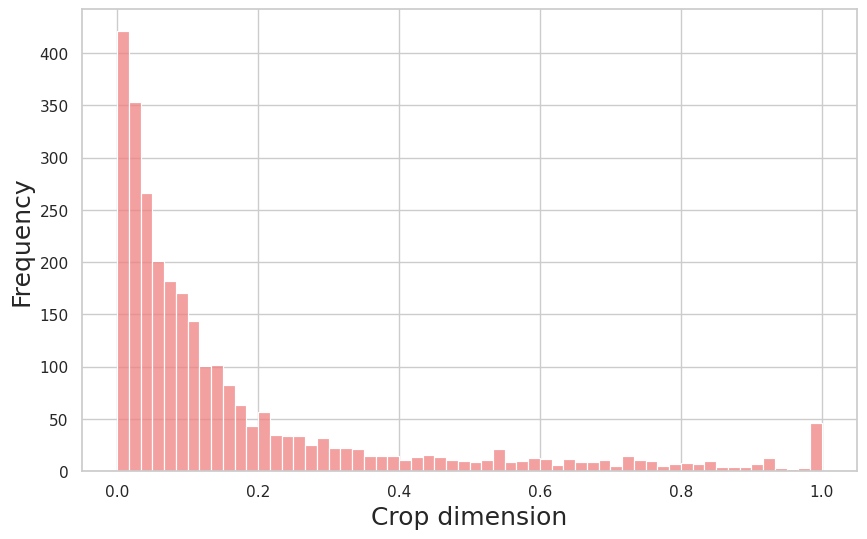}
        \caption{Distribution of crop sizes of objects }
        \label{fig:sub7}
    \end{subfigure}
    \hfill
    \begin{subfigure}[b]{0.48\textwidth}
        \hfill \vspace{0.2cm}
        \includegraphics[width=0.95\textwidth]{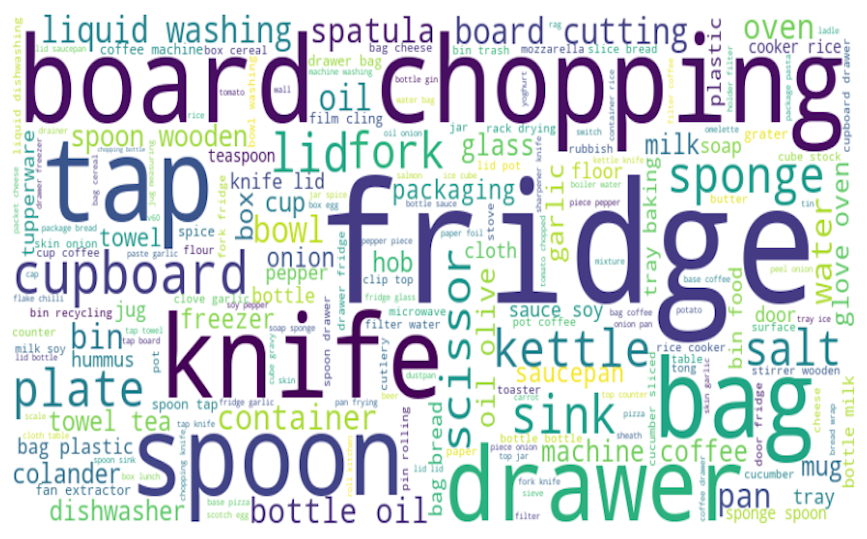}
        \caption{Wordcloud of most frequent queried objects}
        \label{fig:sub8}
    \end{subfigure}
    \caption{Statistics of the \benchmark questions}
    \label{fig:stats}
\end{figure}

To create the \benchmark we randomly sample 100 EPIC-KITCHENS \cite{damen2022rescaling} videos among the ones with both VISOR \cite{darkhalil2022epic} masks and EPIC Fields \cite{tschernezki2024epic} camera poses. 
We use the list of nouns from the narrations available in EPIC-KITCHENS, as an initial set of possible objects.
We then filter out objects without corresponding VISOR masks, as we use these for spatial ground truth.
We then generate the queries for all annotated objects/locations starting from the templates in Table 1 of the main paper. The alternative answers for each query have been generated using a rationale in a semi-automated process to increase the complexity of the benchmark. 

\subsection{Statistics}
In Figure \ref{fig:stats}, various statistics regarding the \benchmark are presented. Specifically, the distribution per query type (Figure \ref{fig:sub1}), per reasoning level (Figure \ref{fig:sub2}), per video duration (Figures \ref{fig:sub3} and \ref{fig:sub4}), the interaction duration for queries of type Q7 (Figure \ref{fig:sub5}) and Q8 (Figure \ref{fig:sub6}), the distribution of crop sizes of the objects (relative to the frame dimension) in our benchmark (Figure \ref{fig:sub7}), and the frequency of queried nouns as a WordCloud (Figure \ref{fig:sub8}). Notably, sequencing queries constitute nearly two-thirds of the entire benchmark, reflecting their importance in understanding the temporal flow of object interactions in long videos, which is essential for higher-level understanding such as causal inference. Moreover, the number of questions increases with longer video durations (\cref{fig:sub4}), resulting in a benchmark tailored towards longer videos (the main focus of our work, see \cref{fig:sub3}). The duration of interactions exhibits a long-tailed distribution due to the fine-grained nature of queried objects and locations (\cref{fig:sub5}) and (\cref{fig:sub6}). Naturally, the most frequent objects in the dataset are also prominent in our questions (\cref{fig:sub8}). Smaller object crops are more frequently involved in our queries (\cref{fig:sub7}).

\subsection{Benchmark comparison}
Compared to other egocentric QA benchmarks, presented in Table~\ref{tab:qa_comparison}, \benchmark stands out due to its unique characteristics. It primarily emphasizes long egocentric videos, evident from the substantial average length of queried recordings. Similar to ReST \cite{yang2023relational}, \benchmark maintains a strong focus on vision, thereby mitigating potential language biases. However, unlike ReST, \benchmark also incorporates the location dimension into its framework.
\begin{table}[ht]
\caption{Comparison of \benchmark with other egocentric video QA datasets}
\label{tab:qa_comparison}
\begin{tabular}{l|cccc}
\toprule
Benchmarks   & \#Queries (K) & Avg. length (s) & Total hours & Vision focused                \\ \midrule
EgoVQA       & 0.5           & 2.2             & 0.3         &                               \\
EgoSchema    & 5             & 180                                  & 253         &                               \\
QAEgo4D      & 14.5          & 495                                  & 182         &                               \\
ReST - ADL   & 185.7         & 1631                                 & 9           & \checkmark (Activity, Object, Time) \\
ReST - Ego4D & 303.3         & 1104                                 & 92          & \checkmark (Activity, Object, Time) \\ \midrule
AMB          & 20.5          & 1207                                 & 22.7        & \checkmark (Location, Object, Time) \\ \bottomrule
\end{tabular}
\end{table}

\section{Answering questions using AMEGO}
\label{sec:query_answer}
Given the memory $\mathcal{E}$, our representation of the long egocentric video, querying it provides various ways to answer questions regarding interacting objects and locations. Initially, we retrieve the closest representation of the query object/location, then apply the obvious logic to address various questions in the \benchmark. Specifically:
\begin{itemize}
    \item \textbf{Q1}: we match all objects in the sequences associated with each answer, assigning each image patch to an object ID $q_{id}$, among those in $\mathcal{O}$. Subsequently, we select the answer with the longest common subsequence calculated between the complete sequence of $\mathcal{O}$ and any of the answers;
    \item \textbf{Q2-3}: we match the query object with the track in $\mathcal{O}$ based on three criteria: (i) temporal proximity to $t$, (ii) containing the hand side specified in the question, and (iii) achieving a minimum similarity of 0.6. Once the matching track is identified, we simply extract the track after (Q2) or before (Q3) with the query hand side in it and search among the answers for the one with the highest similarity;
    \item \textbf{Q4}: we match the query object with the tracks in $\mathcal{O}$ and extract the corresponding object ID $q_{id}$. Using this ID, we identify the first/last location segment where it appeared in $\mathcal{E}$ and compare it with the answers. We select the answer with the highest similarity;
    \item \textbf{Q5}: we match the query object with the tracks in $\mathcal{O}$ and extract the corresponding object ID, $q_{id}$. Similarly, we match each image patch in the answers and extract the corresponding object IDs. Finally, we select the answer with the highest number of object IDs that are concurrent with $q_{id}$ (i.e., overlapping temporal segments) in $\mathcal{E}$; 
    \item \textbf{Q6}: similar to Q5, we match the query object with the tracks in $\mathcal{O}$ and extract the corresponding object ID, $q_{id}$. Then, we match each location patch in the answers and extract the corresponding location IDs. Finally, we select the answer with the highest number of location IDs that are concurrent with $q_{id}$ (i.e., overlapping temporal segments) in $\mathcal{E}$;
    \item \textbf{Q7-8}: we match the query object or location to the tracks in $\mathcal{O}$ or $\mathcal{L}$ respectively. Then, we extract the corresponding instance and retrieve from $\mathcal{E}$ all temporal intervals where the instance was active. Finally, we select the answer with the highest average temporal Intersection over Union (IoU);
\end{itemize}
For each of the cases above, if two or more answers were found to be matching the representation, we select the answer randomly.
Similarly, if no answer is found, a random answer is selected.
We employ straightforward approaches to maintain focus on the strength of the representation rather than the querying method. The potential information extracted from $\mathcal{E}$ is solely constrained by the representation itself, and similar querying techniques can be readily implemented to address diverse, fine-grained questions about interactions in the long egocentric video $\mathcal{V}$.

\begin{table}
\caption{Ablation on IoU value}
\centering
\begin{tabular}{l|c|c|c}
\toprule
IoU & AIoU P $\uparrow$ & AIoU GT $\uparrow$ & $\mathrm{\Delta N} \rightarrow 0$ \\ \midrule
$0.3$ & 0.19 & 0.37 & 226 \\
$0.5$ & 0.20 & 0.41 & 210 \\
$0.7$ & 0.18 & 0.32 & 168 \\
\bottomrule
\end{tabular}%
\label{tab:ablation_iou}
\end{table}
\begin{table*}[t]
\centering
\begin{minipage}{0.45\textwidth}
\begin{tikzpicture}
    \begin{axis}[
        width=\linewidth,
        xlabel={$\theta$},
        ylabel={ID-switch},
        ymin=0.38,
        ymax=0.45,
        xtick={0.4, 0.6, 0.8},
        xticklabels={$0.4$, $0.6$, $0.8$},
        grid=both,
        ]
        
        \addplot[mark=square, blue] coordinates {
            (0.4, 0.44)
            (0.6, 0.41)
            (0.8, 0.43)
        };
    \end{axis}
\end{tikzpicture}

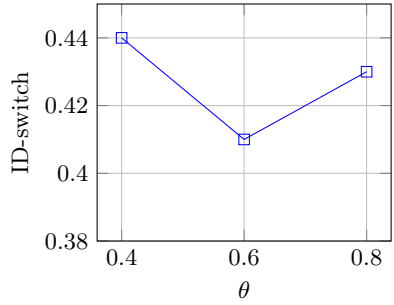
\captionof{figure}{Ablation on $\theta$}
\label{fig:ablation_theta}
\end{minipage}
\hfill
\begin{minipage}{0.45\textwidth}
\begin{tikzpicture}
    \begin{axis}[
        width=\linewidth,
        xlabel={$\tau$},
        ylabel={ID-switch},
        ymin=0.38,
        ymax=0.45,
        xtick={0.3, 0.5, 0.7},
        xticklabels={$0.3$, $0.5$, $0.7$},
        grid=both,
        ]
        
        \addplot[mark=square, blue] coordinates {
            (0.3, 0.44)
            (0.5, 0.40)
            (0.7, 0.42)
        };
    \end{axis}
\end{tikzpicture}

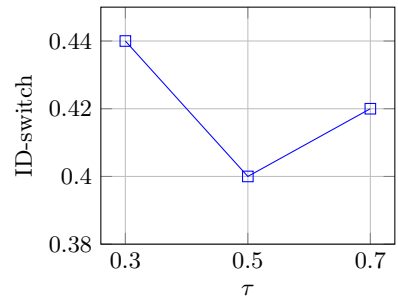
\captionof{figure}{Ablation on $\tau$}
\label{fig:ablation_tau}
\end{minipage}
\end{table*}
%\newpage
\section{Additional ablations}
\label{sec:more_results}
%\subsection{Ablations}

We present here additional ablation results on the clustering threshold to perform the assignment step for both objects ($\theta$) and locations ($\tau$) and for the IoU threshold value adopted for spatial matching of HOI tracklets $\mathcal{O}$ with object detections $\mathcal{B}^o$. We performed the ablations on the two manually annotated videos described in Sec. 5.2 of the main paper. As it is important to evaluate clustering performance to ablate on $\theta$ and $\tau$, we adopt another metric, ID-switch. It computes the average number of times a predicted segment changes its instance consecutively when it should not, i.e. when the ground truth object remains the same. Consequently, we want it to be as low as possible. \cref{fig:ablation_theta} and \cref{fig:ablation_tau} show the effect of the clustering threshold on ID-switch. Although clustering demonstrates stability across various thresholds, our selected threshold proves to be optimal for the two manually annotated videos. Similar deductions can be made from  \cref{tab:ablation_iou}, where it is possible to notice that results do not change much depending on the IoU threshold chosen.

\section{\our pseudocode}
\label{sec:pseudocode}
For the sake of clarity we report here the pseudocode depicting the algorithms to build out interacting objects (\cref{alg:hoi}) and locations (\cref{alg:loc}) representation.
\begin{algorithm}[t]
\caption{Object interactions pipeline}
\label{alg:hoi}
\begin{algorithmic}[1]
\STATE \textbf{Input:} 
\STATE \hspace{1em} Frames $\{\mathcal{V}_t\}$
\STATE \hspace{1em} HOI detector $\mathcal{D}$
\STATE \hspace{1em} SOT tracker $\mathcal{T}$
\STATE \hspace{1em} Similarity threshold $\theta$

\vspace{0.5em}
\STATE \textbf{Output:} 
\STATE \hspace{1em} Set of hand-object interaction tracklets $\mathcal{O}$

\vspace{0.5em}

\vspace{0.5em}
\FOR{each frame $\mathcal{V}_t$}
    \STATE $\mathcal{B}^o_t, \mathcal{B}^h_t \leftarrow \mathcal{D}(\mathcal{V}_t)$ \COMMENT{Detect hands and objects}
    
    \FOR{each detection $(b^o, b^h) \in (\mathcal{B}^o_t, \mathcal{B}^h_t)$}
        \IF{new hand-object interaction (i.e. $s_o$ detections in the last $w_s$ frames)}
            \STATE Create new tracklet $o_i$
            \STATE Start SOT $\mathcal{T}_{o_i}$ for $o_i$
        \ENDIF
    \ENDFOR
    
    \FOR{each tracklet $o_i$}
        \STATE Update the detections with $\mathcal{T}_{o_i}$
        \IF{$\nexists b^o \in \mathcal{B}^o_t$ matching with $o_i$ in the last $e_o$ frames \AND $|\mathcal{B}^h_t| > 0$}
            \STATE Mark $o_i$ as complete
        \ENDIF
    \ENDFOR
    
    \FOR{each completed tracklet $o_i$}
        \STATE Compute visual features $f(o_i)$ (Eqn. 1)
        \STATE Compute similarity $s(o_i, id_j)$ with existing instances in $\mathcal{O}$ (Eqn. 2)
        \IF{maximum similarity $> \theta$}
            \STATE Assign $o_i$ to best matching instance $id_j$
        \ELSE
            \STATE Create new instance for $o_i$
        \ENDIF
        \STATE Store $o_i$ in $\mathcal{O}$
    \ENDFOR
\ENDFOR

\vspace{0.5em}
\STATE \textbf{return} $\mathcal{O}$

\end{algorithmic}
\end{algorithm}
\begin{algorithm}[t]
\caption{Location Segment pipeline}
\label{alg:loc}
\begin{algorithmic}[1]
\STATE \textbf{Input:} 
\STATE \hspace{1em} Frames $\{\mathcal{V}_t\}$
\STATE \hspace{1em} HOI detector $\mathcal{D}$
\STATE \hspace{1em} Similarity threshold $\tau$

\vspace{0.5em}
\STATE \textbf{Output:} Set of location segments $\mathcal{L}$

\vspace{0.5em}

\FOR{each frame $\mathcal{V}_t$}
    \STATE $\mathcal{B}^o_t, \mathcal{B}^h_t \leftarrow \mathcal{D}(\mathcal{V}_t)$ \COMMENT{Detect hands and objects}
    
    \STATE Compute optical flow $\text{OpticalFlow}(\mathcal{V}_{t-1}, \mathcal{V}_t)$

    \IF{location segment $l_j$ is active}
        \IF{high $|\text{OpticalFlow}(\mathcal{V}_{t-1}, \mathcal{V}_t)|$  \OR $|\mathcal{B}^h_t| = 0$ for $e_l$ consecutive frames}
            \STATE Mark $l_j$ as complete
        \ELSE
            \STATE Continue $l_j$
        \ENDIF
    \ELSE
        \IF{low $|\text{OpticalFlow}(\mathcal{V}_{t-1}, \mathcal{V}_t)|$  \AND $|\mathcal{B}^h_t| > 0$ for $s_l$ consecutive frames}
            \STATE Subject is interacting, start active location segment $l_j$
        \ENDIF
    \ENDIF

    \FOR{each completed segment $l_j$}
        \STATE Compute visual features $g(l_j)$
        \STATE Compute similarity $s(l_j, id_i)$ with existing instances in $\mathcal{L}$ 
        \IF{maximum similarity $> \tau$}
            \STATE Assign $l_j$ to best matching instance $id_i$
        \ELSE
            \STATE Create new instance for $l_j$
        \ENDIF
        \STATE Store $l_j$ in $\mathcal{L}$
    \ENDFOR
\ENDFOR

\vspace{0.5em}
\STATE \textbf{return} $\mathcal{L}$

\end{algorithmic}
\end{algorithm}

\clearpage

\end{document}